# A unified deep learning framework for multi-vessel classification, segmentation and phenomapping of phase-contrast MRI validated on a multi-site single ventricle patient cohort

*Tina Yao[†], Nicole St. Clair [†], Madeline Gong, Gabriel F. Miller, Michael Quail, Shahin Moledina, Adam L. Dorfman, Mark A. Fogel, Rajesh Krishnamurthy, Christopher Z. Lam, Joshua D. Robinson, Timothy C. Slesnick, Justin Weigand, Jennifer A. Steeden, Rahul H. Rathod, Vivek Muthurangu* and FORCE Investigators*

** V.M is the corresponding author*

*[†] T.Y. and N.S.C. contributed equally to this work.*

**Abstract**

We present a deep learning framework with two models for automated segmentation and large-scale flow phenotyping in a registry of single ventricle patients.

MultiFlowSeg simultaneously classifies and segments five key vessels—left and right pulmonary arteries, aorta, superior vena cava, and inferior vena cava—from velocity encoded phase-contrast magnetic resonance (PCMR) data. Trained on 260 CMR exams (5 PCMR scans per exam), it achieved an average Dice score of 0.91 on 50 unseen test cases. The method was then integrated into an automated pipeline where it processed over 5,500 registry exams without human assistance – in exams with all 5 vessels it achieved 98% classification and 90% segmentation accuracy.

Flow curves from successful segmentations were used to train MultiFlowDTC, which applied deep temporal clustering to identify distinct flow phenotypes. Survival analysis revealed distinct phenotypes were significantly associated with increased risk of death/transplantation and liver disease, demonstrating the potential of the framework.

## MAIN

Time-varying signals can provide important insights into cardiac pathophysiology, capturing dynamic processes that static measurements cannot fully characterize. For instance, it is well recognized that certain blood flow patterns are associated with specific disease processes (e.g., an abnormal early/atrial filling ratio in diastolic dysfunction)[1]. One of the most accurate ways of measuring time-varying blood flow is velocity-encoded phase-contrast magnetic resonance imaging (PCMR). This technique is heavily used in the evaluation of numerous cardiovascular disorders, including congenital heart disease (CHD)[2–9] and valvular disease. However, usually only time-averaged metrics (e.g. net forward volume) are reported and these do not harness the full potential of time-varying signals. More sophisticated methods of leveraging the time-varying nature of PCMR data have been developed, but access to large amounts of data is required to validate their clinical relevance. Such datasets are becoming more common with the increasing availability of large cardiac magnetic resonance (CMR) registries. Unfortunately, most of these registries only contain clinically reported metrics such as net forward volume and unprocessed PCMR images. Thus, in order to be used to validate/develop more sophisticated methods for time-varying flow analysis, the PCMR images within these registries needs to be reprocessed (segmented). Manual segmentation of most registries is not tractable, being both time-consuming and prone to inter- and intra-observer variability[10]. Deep learning (DL) offers the possibility of both automatically processing registry-held PCMR images and aiding in developing more sophisticated methods of leveraging time-varying flow data. Although such a framework could apply to multiple diseases, we chose to focus on a challenging exemplar – patients with functional single ventricles and a Fontan circulation.

The Fontan circulation is characterized by very complex anatomy, and highly abnormal flow patterns[11–14]. It is thus well suited to an analysis of time-varying flow data and a registry containing over 5,000 Fontan CMR exams from 44 international sites already exists (Fontan Outcomes Registry using CMR Examinations – FORCE, http://www.forceregistry.org)[15]. This includes unprocessed PCMR images of the left and right pulmonary arteries (LPA and RPA), aorta (Ao), and superior and inferior vena cavae (SVC and IVC) in most patients. However, automated segmentation of the PCMR images in this registry presents two significant challenges: (i) The images are very heterogeneous due to both differing CMR protocols/hardware at the contributing sites (common to all registries) and highly variable anatomy (specific to Fontan patients), and (ii) There are multiple distinct vessels that first must be identified and then segmented and if processing is to be done without human input. Our solution is a single DL model (MultiFlowSeg) that uses a UNet3+ architecture (that we have shown works with other images in FORCE registry[16]) to simultaneously classify and segment multiple vessels.

After automatic processing of PCMR images in the FORCE registry, it was possible to develop novel methods to analyze time-varying flow data. We choose clustering, which is an increasingly popular method of interrogating complex data, particularly in large populations. Clustering attempts to group patients based on similar characteristics in some n-dimensional space and has been used to discover new phenotypes (phenomapping) in several cardiovascular diseases[17–19]. Although conventional clustering has been used to analyze time-varying signals (e.g. using dynamic time warping)[20–23]. We chose Deep Temporal Clustering (DTC)[24] that uses DL to improve clustering performance. Combining DL segmentation (MultiFlowSeg) with DL phenomapping (MultiFlowDTC) provides a unique opportunity to investigate the time-varying blood flow in the Fontan circulation, whilst also being easily generalizable to other pathologies.

The specific aims of this study were: (i) To develop and validate a unified DL model capable of classifying and segmenting five different phase-contrast flow planes, (ii) To integrate the DL model into an automated pipeline to process the whole of the FORCE registry data (>5000 datasets) and evaluate segmentation quality, (iii) To use extracted time-varying flow curves from the FORCE registry to perform deep temporal clustering and identify novel flow-based phenotypes, and iv) Associate these phenotypes with key clinical outcomes including: ejection fraction, exercise tolerance, liver disease, and mortality.

# RESULTS

Figure 1 shows the overall architecture of the MultiFlowSeg and MultiFlowDTC modules (full details in methods section). MultiFlowSeg has a modified UNet3+ architecture with full-scale skip connections, deep supervision and a tunable input based on the DICOM series description (entered by the technologist at the time of scanning). MultiFlowSeg was trained, validated and formally tested on manually segmented PCMR images from the FORCE registry (185/25/50 exams each containing five 2D+time datasets). The MultiFlowSeg model was then integrated into a fully automated processing pipeline, with segmentation results rated by a segmentation expert. MultiFlowDTC is based on a temporal autoencoder backbone with additional latent space-based clustering. Unsupervised learning using MultiFlowDTC was performed on the pipeline-processed time-volume flow curves from >5000 exams. The flow phenotypes discovered using MultiFlowDTC were then associated with clinical outcomes.

## MultiFlowSeg Evaluation

MultiFlowSeg was able to successfully process all the test data with an inference time of 218ms per cine on a GPU (NVIDIA RTX A6000). In comparison, manual segmentation with a semi-automated segmentation algorithm took ~30-40s per cine.

### Classification Performance

On the 50 unseen exams (250 PCMR series) used for testing, MultiFlowSeg achieved 100% accuracy in identifying the Ao, SVC, and IVC, and 94% accuracy for the LPA and RPA (Figure 2A).

A key aspect of MultiFlowSeg was the tunable parameter that leverages series descriptions to improve classification performance, whilst being robust to missing or incorrectly entered descriptions. To evaluate the model's flexibility, we additionally compared: (i) The MultiFlowSeg model described above, (ii) A “Vanilla” model without the tunable input, (iii) A MultiFlowSeg model with missing series descriptions, and (iv) A MultiFlowSeg model where all the series descriptions, for every vessel, were artificially set to be incorrect (Figure 2A-D).

In the same 50 test exams, MultiFlowSeg (with actual descriptions) achieved 97% overall classification accuracy, while the vanilla, missing descriptions and incorrect description models achieved 94%, 94% and 90% accuracy, respectively. These results show that the tunable parameter improves classification accuracy, but even when this information is missing or incorrect, MultiFlowSeg maintains high classification performance.

## Segmentation Performance

MultiFlowSeg achieved robust segmentation with a median Dice score of 0.91 (IQR: 0.86–0.93) across all test vessels (n = 250 – Supplementary Figure 1). The Ao and IVC had the highest Dice scores, with medians of 0.93 (IQR: 0.91–0.95) and 0.93 (IQR: 0.89–0.95), outperforming other vessels ($p < 0.007$). The LPA, RPA, and SVC also showed strong accuracy, with median Dice scores of 0.90 (IQR: 0.87–0.93), 0.89 (IQR: 0.85–0.91), and 0.88 (IQR: 0.84–0.92), respectively.

Time-varying flow curves calculated from manual and DL segmentation are shown in Figure 3 demonstrating good agreement over time for all vessels (segmentation examples for the interquartile range Dice for each vessel in Supplementary Figures 2—6). Comparison of net forward volumes calculated from these curves is shown in Figure 4 with clinically acceptable limits of agreement and strong intraclass correlations (ICC>0.95). However, DL segmentation did produce slightly higher net forward volumes than manual segmentations (2.2–6.7%) which were significant for all vessels except the Ao (p=0.06).

## Pipeline Performance

MultiFlowSeg was integrated into an automated pipeline to extract, classify and segment PCMR images from all CMR exams in the FORCE registry. Using this pipeline (see Methods), we automatically processed 5515 full CMR exams on a cloud instance (8 vCPUs, 120 GB RAM). Total processing time averaged 127 ± 80 seconds per exam. This time included: i) sorting through all the series in an exam (~31 series per exam), ii) extracting the PCMR series (~11 series per exam), iii) running MultiFlowSeg on all PCMR series (2.9s per series on the cloud instance), iv) storing flow curves and segmentation masks for the LPA, RPA, Ao, SVC and IVC, and v) creation of a GIF showing segmentation results for quality assurance. The total time to process the whole registry was 194 hours, with no human interaction required during this time.

Out of the 5515 processed exams, 5046 included phase-contrast MRI (PCMR) data from at least one vessel. Excluding the 260 exams used for ground truth manual segmentation, 2902 exams had usable PCMR data for all five vessels. The segmentation for the exams with all five vessels were manually reviewed by the researcher who created the ground truth dataset, and rated each vessel segmentation as ‘acceptable’, ‘not acceptable’, or ‘misclassified’ (Supplementary Figure 8).

Of the 2902 exams reviewed, the pipeline achieved an overall 98% classification accuracy. The classification accuracy for the Ao, SVC and IVC was 100%, whereas the LPA and RPA had 96% and 94% classification accuracy, respectively.

The pipeline achieved an overall ‘acceptable’ segmentation rate of 90% across all vessels, with individual success rates as follows: LPA – 92%, RPA – 82%, Ao – 83%, SVC – 96%, IVC – 97%.

Factors that affected segmentation success rate are shown in Supplementary Table 1 and included: (i) Poor image quality (RPA and SVC – $p < 0.03$), (ii) Similarly sized neo and native aortas (LPA, RPA, Ao and SVC – $p < 0.047$), (iii) Bilateral SVC's (LPA, RPA and SVC – $p < 0.004$), and (iv) Heterotaxy (all vessels except the aorta – $p < 0.001$).

## Phenomapping using MultiFlowDTC

We used MultiFlowDTC to perform temporal cluster analysis on flow curves derived from MultiFlowSeg. We trained two clustering models, one combining the LPA and RPA flow curves and the other combining the SVC and IVC flow curves. However, only exams in which the aortic segmentation was acceptable were used data, as the aortic flow curve was used to further remove exams that were pulse gated. In addition, for each of the models' exams were only included if both vessel segmentations were rated acceptable resulting in 1943 LPA/RPA curves, and 2286 SVC/IVC curves.

### Clustering

The optimal number of clusters was determined using the temporal silhouette score, resulting in five clusters for the LPA/RPA clustering model (silhouette score = 0.89) and four clusters for the SVC/IVC model (silhouette score = 0.90). Figure 5 presents t-distributed stochastic neighbor embedding (t-SNE) plots of the clusters before and after joint optimization, demonstrating that the DTC method generates more distinct clusters compared to simple K-means on the latent space (similar to PCA-based methods)[25].

Pulmonary artery (PA) clusters – Figure 6 shows the median flow curves patients in the five PA clusters. Supplementary Table 2 presents patient demographics for the PA model clusters, while Supplementary Table 3 outlines the statistical differences. The PA cluster characteristics are summarized as follows:

1. Normal Distribution, High Flow ($PA_{Norm\text{-}High}$): This group shows a normal distribution of branch PA flow (42% to LPA) and with high total pulmonary blood flow (3.0 L/min/m$^2$). This cohort is the youngest (mean age 13.7 years, 63% pediatric), significantly younger than all other groups except RPA-Norm ($p < 0.001$). They also have the highest heart rate and indexed aortic flow rate ($p < 0.001$).
2. Balanced Distribution, Low Flow ($PA_{Bal\text{-}Low}$): While the branch flow distribution is symmetric, this group has reduced overall total pulmonary blood flow (1.4 L/min/m$^2$), higher aorto-pulmonary collateral flow, the lowest ejection fraction ($p < 0.03$). This group also had a higher proportion of patients with heterotaxy, which was statistically significant compared to $PA_{Norm\text{-}High}$ and $PA_{RPA\text{-}Norm}$ ($p<0.03$).
3. Diastolic-Dominant, Normal Flow ($PA_{Dia\text{-}Norm}$): This group is defined by a greater proportion of flow occurring during diastole, has low/normal total pulmonary blood flow (1.9 L/min/m$^2$) and includes older patients with lower heart rates compared to $PA_{Norm\text{-}High}$ and $PA_{RPA\text{-}Norm}$ groups ($p < 0.001$).

4. RPA-Dominant, Normal Flow ($PA_{RPA\text{-}Norm}$): Characterized by the lowest proportion of flow directed to the LPA (30%, $p < 0.001$ vs. all). It also has a greater proportion of patients with hypoplastic left heart syndrome and lower proportion with heterotaxy than $PA_{Bal\text{-}Low}$ and $PA_{Bal\text{-}Norm}$ ($p<0.022$).
5. Balanced Distribution, Normal Flow ($PA_{Bal\text{-}Norm}$): This group shows balanced flow (52% to the LPA, $p < 0.001$ vs. all) and has a higher proportion of patients with heterotaxy compared to all groups except $PA_{Bal\text{-}Low}$ ($p < 0.01$).

Vena caval (VC) clusters - Figure 6 shows the median flow curves patients in the four VC clusters. Supplementary Table 4 shows the patient demographics for the clusters generated by the VC model, while Supplementary Table 5 presents the statistical differences. The characteristics of the VC clusters are summarized as follows:

1. IVC-Dominant, Normal Flow ($VC_{IVC\text{-}Norm}$): This group has a higher proportion of flow from the IVC (72% to IVC) and is the oldest (mean age 20.6 years, 81% adult, $p < 0.001$). It also has the highest proportion of Lateral Tunnel Fontan procedures ($p < 0.001$) and the lowest exercise tolerance which was significant compared to $VC_{SVC\text{-}High}$ and $VC_{Norm\text{-}High}$ ($p<0.002$).
2. SVC-Dominant, High Flow ($VC_{SVC\text{-}High}$): The only group with more flow directed to the SVC (56%), this is the youngest group (mean age 8.2 years, 95% pediatric, $p < 0.001$). It has the highest proportion of extracardiac conduit Fontan procedures and the highest aortic flow rate ($p < 0.001$). In addition, this group has the highest exercise tolerance, which was significant compared to $VC_{IVC\text{-}Norm}$ and $VC_{Norm\text{-}Norm}$ ($p < 0.001$).
3. Normal Distribution, High Flow ($VC_{Norm\text{-}High}$): With normal flow distribution (37% to SVC), this group is predominantly younger (72% pediatric, $p < 0.001$), though not as young as $VC_{SVC\text{-}High}$. It also has the least collateral flow ($p < 0.001$).
4. Normal Distribution, Normal Flow ($VC_{Norm\text{-}Norm}$): This group also has normal flow distribution (37% to SVC) with the lowest ejection fraction and aortic flow rate ($p < 0.001$ vs. all except $VC_{IVC\text{-}Norm}$).

### Cluster Outcome Analysis

Given that patients have multiple scans and their cluster membership can change over time, we performed time-varying covariate time-to-event analysis (time-varying Cox regression[26] and Simon-Makuch plots - an extension of the Kaplan-Meier method for time-varying groups[27,28]). We investigated two outcomes (i) Death or transplantation, and (ii) The first diagnosis of Fontan associated liver disease.

Figure 6 shows the Simon-Makuch plots for death/transplantation and liver disease for both the PA and VC clusters. Hazard ratios comparing all clusters are shown in Supplemental Figure 9.

The $PA_{Bal\text{-}Norm}$ group had a significantly lower risk of transplant or death compared to those in the $PA_{Bal\text{-}Low}$ group (HR=3.17, p=0.005). The $VC_{SVC\text{-}High}$ (greater SVC flow) had significantly lower risk death/transplant compared to the $VC_{Norm\text{-}Norm}$ group (HR = 2.56, p=0.033) and the VCIVC-Norm (greater IVC flow) group (HR=x, p = y). The $VC_{SVC\text{-}High}$ group also had a significantly lower risk of liver disease compared to the $VC_{Norm\text{-}Norm}$ group (HR=1.74, p=0.021) and the $VC_{IVC\text{-}Norm}$ (greater IVC flow) group (HR=1.67, p = 0.031).

## DISCUSSION

To our knowledge, this is the first study to develop a unified deep learning model for classification and segmentation of multiple vessels (imaged using PCMR) and flow-based phenomapping of Fontan patients. The key findings were: (i) MultiFlowSeg achieved high accuracy in both classification and segmentation of the major blood vessels, (ii) The integration of MultiFlowSeg into an automated pipeline enabled efficient processing of the entire FORCE dataset (>5000 exams), delivering reliable flow quantification in approximately 90% of the registry, (iii) Deep temporal clustering identified distinct patient clusters based on flow patterns, and (iv) The clusters were significantly associated with key clinical outcomes, including liver disease and death/transplantation. Our unified approach enables the full potential of PCMR images stored in the FORCE registry to be realized, identifying novel physiologically distinct groups with prognostic significance. Importantly, our approach could be easily adapted to other CMR registries that contain PCMR images (e.g. Indicator Cohort or PVDOMICS[29,30]).

### Deep learning PCMR segmentation

Manual core-lab segmentation of the PCMR images in the FORCE registry (>5000 exams) would take ~1300 person hours (~30 working weeks of continuous segmentation). This is not realistically achievable, which is why we developed a DL-based processing pipeline. Our pipeline processed the entire registry without human interaction in 194 hours (even when run on a cloud CPU), significantly reducing resource usage. However, efficient processing is only valuable if classification and segmentation are robust. We demonstrated high Dice scores and strong agreement between flow curves from manual and DL segmentation, despite the challenging and complex anatomy of Fontan patients[31]. We believe this is primarily due to the underlying UNet3+ architecture[32] which allows better multi-scale feature fusion and provides a richer understanding of both fine details and global context. Although we previously used a UNet3+ to successfully segment short-axis images in the FORCE registry[16], significant modifications were made to optimize the architecture for flow segmentation. Firstly, the network was modified to accept 2D+time images by using 3D convolutions. This enabled information flow between consecutive frames, enforcing temporal consistency and preventing segmentations from dropping out or excessively shifting between frames. Secondly, instead of just using the magnitude images, the imaginary component was also used as an input to our model. This enabled velocity information to be leveraged to improve accuracy without having to manage the high noise present in the original phase images. Finally, one of the most important differences between our approach and previous studies was the use of a single network to perform both segmentation and classification. Previous studies segmented multiple vessels using separate classifiers and vessel-specific networks, adding complexity to training and inference while limiting information sharing between tasks[33]. In contrast, our simpler joint approach leverages the UNet3+ architecture and introduces a novel tunable input based on series descriptions. This input provides contextual information added by technologists during CMR scans without hard coding, thereby improving classification accuracy while remaining robust to missing (~7% of all exams) or incorrect descriptions.

Our models demonstrated robust performance across various scanners, clinical sites and complex anatomies, which was vital for our application. When applied to the whole registry, our model reaches 90% success, providing vast amounts of data for flow-based phenotype discovery. Nevertheless, there were certain situations in which MultiFlowSeg performed less well. Some were expected, such as poor image quality, but others were due to less common anatomical variations. For instance, abnormal situs can confuse the model because the LPA can look like the normal RPA plane and vice versa. Other anatomical variations, such as bilateral SVC and similarly sized neo- and native aortas can also impact model accuracy. These cases were underrepresented in the training data, and future performance could improve by including more examples of such uncommon anatomies.

## Temporal Clustering Analysis

Deriving clinically useful biomarkers from time-varying data requires temporal dimensionality reduction. This is often achieved using very simple methods (such as reporting peak or mean values), and for PCMR images, usually only net forward volumes are reported[34,35]. However, these simplistic approaches don't extract all the potentially useful physiological information from flow data. An example of a more sophisticated method of interrogating flow curves is temporal principal component analysis (PCA). Here, flow curves are broken down into the weighted sum of specific flow motifs (principal components – PCs) that account for the majority of variance in the population. The subject specific weights can then be used as biomarkers and temporal PCA analysis has been performed in the Fontan circulation with certain flow patterns (diastolic dominant PA flow) being associated with poorer outcome[36–38]. However, PCA-based methods have several disadvantages including: (i) Decomposition is constrained to linear summation of orthogonal PC's, (ii) Most of the population variance is explained by the first few components, leading to over interpretation of lower order PCs, and (iii) PCs are statistical, not physiological, resulting in potential incorrect interpretation when approached as independent markers. An alternative approach is temporal clustering using dynamic time warping or temporal PCA followed by conventional clustering[39–42]. The benefit of these clustering approaches is that subjects are grouped within a n-dimensional space (defined by time points or temporal PC's) that takes the totality of time-varying data into consideration. However, optimum clustering benefits from approximately 'lossless' compression (not guaranteed using PCA) and highly separable latent spaces (not always the case with dynamic time warping). Thus, we used deep temporal clustering, a method that has been shown to provide superior clustering. In DTC, a temporal autoencoder is used to efficiently perform less lossy non-linear dimensionality reduction. This latent space is further optimized by a temporal clustering layer that yields more distinct latent spaces and higher-purity clusters, as seen in our results. However, it should be noted that DTC works best with large datasets. Thus, the ability to interrogate large datasets is vital, further strengthening the case for our joint segmentation and unsupervised clustering framework. In fact, our clusters are based on > 20x more data than previous studies on blood flow in Fontan patients[36–38].

Deep temporal clustering analysis of the FORCE dataset revealed distinct flow patterns that in some cases were associated with clinically relevant outcomes. Patients in the $PA_{Bal\text{-}Low}$ grou (balanced distribution, low flow) had an increased risk of death. This might be unsurprising, but our analysis was corrected for indexed aortic flow rate and EF, suggesting that this was not simply the result of poor perfusion and cardiac function. Interestingly, this group did have the highest systemic to pulmonary collateral flow and one possibility is that they have higher pulmonary vascular resistance, which would explain their higher mortality. It is also possible that low pulsatility and diastolic dominance in these patients reflect an additional adverse physiology that contributes to higher mortality. A more surprising finding is that the $PA_{Bal\text{-}Norm}$ group with the best outcome has slightly greater flow to the left lung. This slight skewing of blood flow to the left side was not in keeping with the left dominance due to RPA stenosis or heterotaxy (although heterotaxy was slightly overrepresented). Furthermore, the better outcome in these patients would go against the idea that was true maldistribution. An intriguing explanation for these findings is that Fontan anatomy with equal or slightly greater left lung flow is associated with better hemodynamics and thus, improved outcome. This could be investigated by leveraging DL-based anatomical segmentation and flow field estimation[43] to investigate power loss and TCPC resistance[44].

Another finding is that patients in the $VC_{Norm\text{-}Norm}$ group (normal distribution and normal flow) also have increased risk of death/transplantation. This group do have a higher proportion of low flow PA flow subjects and so this finding is in keeping with our findings in the pulmonary arteries. Conversely, patients with dominant SVC flow were associated with a lower risk of death/transplantation and liver disease. These patients were younger, which is expected as SVC/IVC flow ratio reduces with age. Nevertheless, the association with outcome remained even after correction for age, suggesting that other mechanisms resulted in better prognosis in these patients. One possibility is that increased splanchnic circulation is associated with worse mortality This idea is backed up by the fact that the cluster with dominant IVC flow also had higher mortality and a potential mechanism is increased liver disease in both these groups. Another possible explanation is that higher SVC flow leads to less flow collision between the vena caval flow streams, resulting in great hemodynamic efficiency. Further investigation of flow distribution is required to better understand the reason for these associations, but this is very tractable using the large amounts of comprehensive data available in the FORCE registry.

We have demonstrated the potential of deep temporal clustering for the analysis of time-varying signals at scale. This method could easily be applied to multiple time-varying signals in the cardiovascular imaging space including flow in other pathologies, myocardial strain and ventricular/atrial volumetric curves. However, use in large registries does require a joint segmentation-clustering approach.

## Limitations

The main limitation of our study was that segmentation was not successful in all cases. Although this meant that a human did need to review all the outputs of MultiFlowSeg, this can be done in a matter of seconds, which is feasible even when reviewing thousands of exams. Furthermore, unsuccessful segmentations resulted in a reduction in the number of flow curves available for clustering. This could result in selection bias, and therefore improving segmentation accuracy in the future would be beneficial. One solution is to use the several thousand good quality segmentations we now possess to retrain new more sophisticated DL models. However, the complex anatomy of the Fontan circulation may lead to a ceiling in terms of potential segmentation accuracy.

Another issue is that our tunable input relies on a curated data dictionary to extract tokens from series descriptions. This dictionary was developed using descriptions from the FORCE registry, so it may not perform well with datasets using different naming conventions. This could be remedied by using a large language model (LLM) to encode the series descriptions, with end-to-end training of both the UNet3+ and LLM. However, we demonstrated that our model does have high classification accuracy and an LLM may be an unnecessary complication.

Furthermore, we have not demonstrated that our single model performs better than an approach that uses separate classification and segmentation networks. Nevertheless, even if our approach is not more accurate, it is much simpler to train and deploy.

Finally, we were unable to assess the effect of different types of acquisition (e.g. free breathing, breath-hold and real-time) and known artefacts (e.g. due to stents, or background phase) on both segmentation accuracy and cluster assignment. This is because this data wasn't available in the FORCE registry, but the high success rate in segmentation does suggest good overall generalizability.

## Conclusion

We introduced a unified framework for segmentation and clustering. Our segmentation model, MultiFlowSeg, processes multiple blood vessels using phase-contrast images from different sites, scanners, and cases of single-ventricle physiology—a complex congenital heart disease with significant variability among patients who have undergone different surgeries. We validated our segmentation model on a large dataset (2902 samples), achieving an average success rate of 90% across all five vessels. By leveraging the flow curve data, we used MultiFlowDTC to identify clusters and analyze curve characteristics, linking them to clinical outcomes. We believe our method can provide new insights into Fontan physiology and potentially provide new methods of treatment. In addition, our methodology could easily be modified for other large cardiovascular datasets that contain time varying signals (e.g. PCMR, strain data or even ECG data).

# METHODS

This was a multicenter study approved by the Institutional Review Boards or research ethics committees at each participating institution or via a reliance Institutional Review Board agreement with Boston Children's Hospital. The study proposal and this manuscript were approved by the FORCE Data Governance and Publications Committee. The complete list of FORCE Investigator co-authors and affiliations is enumerated in Supplemental Table 1. The code for the implementation of MultiFlowSeg and MultiFlowDTC has been made publicly available: https://github.com/Ti-Yao/MultiFlow.

## MultiFlowSeg

### Model Architecture

MultiFlowSeg (Figure 1) is based on a five-scale UNet architecture with increasing filters at each scale (16, 32, 64, 128 and 255). An additional tunable layer (1 filter) is concatenated at the bottleneck to increase the total number of filters at this scale to 256. Modifications implemented to optimize the model for simultaneous classification and segmentation of multiple vessels and are described below.

Dual Channel Input: PCMR images consist of both magnitude (anatomical) and phase (velocity) data. Conventionally, only magnitude images are used for segmentation, but blood flow can also be useful for vessel identification. However, phase images have high-velocity noise in areas of low signal, which can make learning difficult. Thus, we converted the magnitude and phase images to a complex representation and extracted the imaginary component where flowing blood has a high signal, and static tissue has a near-zero signal (Supplementary Movie 1). Both the magnitude and the associated imaginary images were used as inputs in our model.

Full-scale Skip Connections and Deep Supervision: Inspired by the UNet3+, we incorporated full-scale skip connections and deep supervision into our UNet[32]. These modifications integrate coarse and fine-grained features at multiple scales, and we have previously shown that this architecture is well-suited to segmentation of complex and heterogeneous anatomy[16].

Incorporating Time: The MultiFlowSeg architecture utilizes 3D convolutional kernels that can process the 2D + time PCMR cine images in one pass. The 3D convolutions result in information between frames and produce more consistent segmentation across time. It should be noted that no maxpooling is performed in time, as this would result in unwanted temporal compression.

Tunable Series Description Input: In a tunable UNet[45], additional information is transformed into matrix that can be concatenated to the bottle neck feature maps. This information then propagates up through the decoding branch, aiding the application task. We adapt this idea by incorporating information from the DICOM series description (which is inputted by the technologist at the time of scanning). Specifically, the Tunable input is a one-hot encoded representation of the image plane derived from the series description (that usually describes the image plane). This encoding is generated using a data dictionary containing predefined terms commonly associated with the five vessels (Supplementary Table 7 – Tunable Input Data Dictionary). The one-hot encoding comprised of six elements: one for each vessel (LPA, RPA, SVC, IVC and Ao) and one for instances with empty series descriptions or terms that were missing from the data dictionary. The one-hot encoded series description is processed through the multi-layered perceptron (MLP) and then reshaped to (8x8x1) then tiled to size (8x8x32x1) to match the size of the UNet bottleneck feature map (8x8x32x255). The reshaped features are concatenated with the bottleneck features to yield (8x8x32x256), enabling the tunable information to propagate through the decoding arm.

<u>Multiclass Classification:</u> We incorporated a multichannel output in which each channel was dedicated to producing a specific vessel segmentation mask, enabling joint classification and segmentation. We augmented the multiclass classification by also including a Classification-Guided Module (CGM) that was also inspired by the UNet3+[32]. The CGM was an additional branch from the bottleneck layer that attempts to constrain segmentations to a single output channel. The multiclass classification module in MultiFlowSeg generates a six-class classification output: one for each vessel and one for the background.

**Training Data**

The training dataset consisted of 260 CMR exams (with PCMR images for the LPA, RPA, SVC, IVC and Ao) of single ventricle patients from the FORCE registry: 185 for training, 25 for validation, and 50 for testing. The training set maintained a similar site distribution to the FORCE registry database, while the validation and test datasets had approximately equal numbers from each site, for non-biased model evaluation across different sites. The training data came from 15 sites, while the validation and test data were from 18 sites to better assess generalizability. Supplementary Table 7 illustrates the demographic information of the ground truth data. No significant differences in BSA, age, sex, situs type, heterotaxy, diagnosis, or dominant ventricle were found between the training, validation, and test datasets, though the validation set had fewer patients with lateral tunnel surgeries than the training set ($p = 0.008$).

A clinical researcher with five years of cardiac imaging experience identified phase-contrast exams for each of the five vessels, segmenting only the vessel of interest for each view, even if other vessels were visible (e.g., the aorta in an SVC-specific plane). The vessels were contoured over the entire cardiac cycle using a semi-automatic algorithm with manual correction (Circle cvi42 - version 6.1.2; Circle Cardiovascular Imaging).

### Pre-processing, Model Training, and Post-processing

Data preprocessing was performed to create consistently sized images. First the pixel size was reduced to 2x2mm spatial resolution through spline interpolation, and the resultant images were center cropped or square padded to a size of 128x128 pixels. Reducing the image size helped to keep memory requirements down during training and had no discernable effect on final accuracy. The number of time frames was interpolated to 32 (median for all exams) using spline interpolation. Contrast-limited adaptive histogram equalization (CLAHE) was applied to the magnitude images to improve contrast and generalizability[46]. The imaginary images were calculated from the CLAHE-adjusted magnitude and the phase images.

MultiFlowSeg was trained using a weighted combination of focal Tversky loss for segmentation and categorical cross-entropy for the classification-guided module, with respective weights of 1.00 and 0.25. The combined loss was calculated at each deep supervision layer, with the first four layers equally weighted at 0.25 and the final layer at 1.00. The model was trained for 400 epochs with a batch size of 8 using the Adam optimizer, and only the epoch with the lowest validation loss was saved. On-the-fly image augmentation included random adjustments to brightness, contrast, flipping, rotation, cropping, padding and resizing to improve segmentation generalizability. Additionally, the one-hot encoded series description was randomly altered 5% of the time to help the model handle incorrect or missing descriptions. To address inconsistent flow encoding, the sign of imaginary image pixel values was alternately flipped during training, making the model agnostic to flow direction.

For post-processing, we removed unconnected components from segmentation masks that did not persist across all time frames. For the LPA, RPA, and IVC masks, only the largest connected component was retained. For the aortic and SVC masks, the two largest components were kept to account for neo and native aortas or bilateral SVCs. The predicted segmentation masks were then resized and projected onto the original PCMR images. Flow curves were then generated from both DL and manual segmentation as previously described[47], and net forward volumes were calculated by integrating the flow curve over time.

### MultiFlowSeg Evaluation

MultiFlowSeg was validated on five flow planes from the 50 ground truth test datasets. The classification accuracy of our model was evaluated using a confusion matrix, with accuracy assessed per vessel. We tested the MultiFlowSeg model under different conditions by altering the input series description in the Tunable layer: (i) The "Actual" description (original description), (ii) The "Missing" description (empty description), and (iii) The "Incorrect" description (random incorrect description). These were compared to a "Vanilla" model, which was trained identically to MultiFlowSeg but lacked the Tunable series description input layer.

Segmentation accuracy of our model was evaluated by calculating the Dice score between the DL model predictions and the ground truth. Derived flow curves and net forward volumes are compared between ground truth and predictions for Dice scores greater than 0.

### Pipeline Overview

We developed an automated pipeline that automatically extracts phase-contrast images from the FORCE registry. Supplementary Table 7 illustrates the demographic information of the data which was tested through the pipeline.

The first step in the pipeline is to identify all phase images using DICOM header information (non-zero velocity-encoding value (VENC) with a minimum of 20 time points). The paired magnitude images were then identified by matching orientation (Image Orientation Patient Attribute), position (Image Position Patient Attribute), and ensuring that the image creation times (Instance Creation Time Attribute) were within five minutes. This approach was robust across different scanner vendors that store phase-contrast images with different protocols. All 2D+time image blocks were subsequently preprocessed and inputted in MultiFlowSeg as described above.

It should be noted that a patient exam may contain more PCMR series than the five vessel planes on which the model was trained. MultiFlowSeg will attempt to classify these series into one of our five categories, potentially resulting in misclassification (e.g. left pulmonary vein misclassified as the left pulmonary artery or a scan plane). If multiple series are classified as the same vessel, and only one has a series description that is consistent with the model classification then that one is chosen (e.g., if the classification is 'LPA' and the description includes 'LPA'; see Supplementary Table – 7 Reference Data Dictionary). If this is not the case, we pick the one with the highest probabilities from the classification-guided module.

Once all vessels had been identified, flow curves and net forward volumes were calculated as described above and for quality assurance GIFs of the segmented images were generated. To evaluate pipeline robustness, a single researcher reviewed each exam, categorizing vessel segmentation as ‘acceptable’, ‘not acceptable’, or categorizing the vessel as ‘misclassified’. Additionally, exams were evaluated for image quality, the presence of bilateral aortas and SVCs, and situs type for each exam.

## MultiFlowDTC

We used an implementation of the Deep Temporal Clustering model[24], a fully unsupervised temporal clustering framework to cluster the extracted time-varying flow curves. The model architecture features a temporal autoencoder and a temporal clustering layer (Figure 1).

### Temporal Autoencoder

The temporal autoencoder was used to create compact latent representations ($z_i$) from sequential data. The encoder consisted of a 1D convolutional layer (50 filters, kernel size = 10), max pooling (pool size = 3), and two bidirectional long short-term memory layers (50 and 1 units). The decoder reconstructed sequences using a time-distributed fully connected layer (50 filters), followed by upsampling and a deconvolutional layer (kernel size = 10). Our temporal autoencoder featured a time-distributed fully connected layer which was not present in the original paper[24] but was used to provide more accurate reconstructions of the input data[48].

### Temporal Clustering Layer

The clustering layer grouped similar temporal patterns in the latent space and was first described in the Deep Embedded Clustering method[49]. It consisted of k centroids initialized via K-means on the autoencoder's latent representations. Cluster assignments were computed using Student's t-distribution, which measured similarity between feature vectors and centroids based on Euclidean distance. The soft cluster assignments $q_{ij}$ represent the probability that the data point $z_i$ (the latent feature vector) belonged to cluster j. These were computed using Student's t-distribution, which is defined as:

$$q_{ij} = \frac{\left(1 + \left|\left|z_i - \mu_j\right|\right|^2 / \alpha\right)^{\frac{-(\alpha+1)}{2}}}{\Sigma_{j'}\left(1 + \left|\left|z_i - \mu_{j'}\right|\right|^2 / \alpha\right)^{\frac{-(\alpha+1)}{2}}}$$

where $z_i$ is the latent feature vector for data point i, $\mu_j$ is the centroid of cluster j, α is a hyperparameter that controls the sharpness of the distribution (set to 1). The denominator was a normalization term to ensure that $q_{ij}$ summed to 1 across all clusters for each data point.

### Minimization of KL Divergence

Xie et al. defined the target distribution[49]:

$$p_{ij} = \frac{\left(q_{ij}^2 / \Sigma_i q_{ij}\right)}{\Sigma_{j'}\left(q_{ij'}^2 / \Sigma_i q_{ij'}\right)}$$

where $q_{ij}$ was the soft assignment of data point i to cluster j, the numerator squares $q_{ij}$ to increase the confidence of high-probability assignments, the denominator normalizes the target distribution so that $p_{ij}$ remains a valid probability distribution.

By minimizing the KL divergence between the soft assignments $q_{ij}$ and the target distribution $p_{ij}$ the model encourages the soft cluster assignments to become more similar to the target distribution, resulting in higher-confidence cluster assignments.

### Joint Optimization

To establish meaningful latent representations before clustering, the autoencoder was pretrained for 100 epochs with a learning rate of 0.001. The initial latent space of the autoencoder was used to initialize cluster centroids in the temporal clustering layer to create soft cluster labels. The autoencoder and clustering layer were then optimized jointly using the Adam optimizer (learning rate = 0.001), minimizing both mean squared error for accurate latent space representation and the Kullback-Leibler (KL) divergence to increase cluster clarity[50], with both losses equally weighted. Cluster assignments were updated every 100 epochs until convergence was reached, defined as fewer than 0.1% of samples changing cluster assignments.

We trained two MultiFlowDTC models for the branch pulmonary arteries, and inferior and superior vena cavae. To determine the optimal number of clusters, each model was trained with 3–7 clusters and the one that achieved the highest temporal silhouette score (ranging from -1 to +1) was chosen[51]. Both models shared the same architecture and training parameters. As MultiFlowDTC is fully unsupervised, the entire dataset was used for both training and inference.

### Preprocessing Flow Curve Data

Flow curve data were included only with acceptable aortic segmentation to check for correct cardiac gating—either from manually segmented (N = 260) or rated ‘acceptable’ from the automated pipeline (N = 2396). Patients whose aortic flow curves did not peak in the first half of the cardiac cycle were excluded, as this indicated incorrect cardiac gating during imaging (N = 21) and any exams with aortic flow that seemed non-physiological (<10mL) were removed (N = 7). We also excluded exams with missing nominal interval (N = 16) and BSA information (N = 196) which were necessary for calculating BSA-indexed flow curves from the segmentations.

After exclusions, 2416 exams had acceptable aortic segmentations and complete demographic data; of these, 1943 had acceptable LPA/RPA and 2286 had acceptable SVC/IVC segmentations.

Because PCMR scans are cardiac-gated, the flow curves had varying time scales depending on the patient's nominal interval at the time of scanning. However, since each curve represents a single cardiac cycle, they remain directly comparable. To have consistent data, all flow curves were BSA-indexed and cubic-spline interpolated to 30 frames, which was the median number of images in a PCMR series. The flow curve values were then standardized by subtracting the mean of the data and dividing by the standard deviation. Processed flow curves from the LPA/RPA or SVC/IVC were then concatenated and treated as separate channels. The input shape was defined by the number of sequences, the length of the time series, and the vessel channels.

**Cluster Outcome Analysis**

To evaluate the prognostic significance of flow phenotypes derived from our MultiFlowDTC models, we assessed their association with two clinical outcomes: (i) heart transplantation or death, whichever occurred first, with transplantation considered equivalent to death due to its role as a life-saving intervention and (ii) first diagnosis of Fontan associated liver disease. The number of events for each outcome for the datasets used in the two models is presented in Supplementary Table 8.

Flow curve clusters were assigned independently for each PCMR exam. Since many patients underwent multiple exams, these cluster assignments were modeled as time-varying covariates to capture temporal changes in vascular flow patterns. Time-dependent Cox proportional hazards models were used to estimate risk, adjusting for age, sex, body surface area (BSA) indexed aortic flow rate and ejection fraction to control for potential confounders. Each cluster was used once as the reference group when calculating hazard ratios.

Risk was visualized using Simon-Makuch plots, which extend Kaplan-Meier curves to incorporate time-varying covariates. Plots were truncated when fewer than ten individuals remained at risk in any cluster to preserve interpretability.

To explore potential explanations for differences in survival, we compared patient characteristics and clinical metrics across the identified flow phenotypes. Patient characteristics included age, sex, body surface area, type of Fontan procedure, type of systemic ventricular circulation, presence of heterotaxy, situs type, and diagnosis. Clinical metrics obtained from the same session as the PCMR included end-diastolic volume, end-systolic volume, and ejection fraction, all derived from short-axis cine imaging. PCMR-specific measures included collateral flow, the distribution of flow across vessels, and the flow rate in each of five key vessels. Additionally, exercise tolerance (Percentage of Predicted $VO_2$), obtained from exercise stress testing, and end-diastolic pressure (EDP), measured via cardiac catheterization, were included when available. These values were used if collected within one year before or after the corresponding CMR.

## Statistical Analysis

Continuous variables are expressed as medians with IQRs, as most variables were not normally distributed.

Kruskal-Wallis and post-hoc Dunn (Benjamini/Hochberg correction) tests were used to find statistical differences in Dice score between vessels. Bland-Altman and intraclass correlation were used to assess agreement between net forward volumes derived from DL and manual segmentations. Wilcoxon signed-rank tests were used to assess the significance in differences between the manual and DL flow calculations. Spearman's correlation coefficient was used to measure the correlation between the Dice score and the percentage difference in segmentation-derived net forward volume between the DL and manual segmentations.

To compare patient demographics across train, validation, and test ground truth datasets, as well as between patient clusters identified by the two MultiFlowDTC models, we used the Kruskal-Wallis test with post-hoc Dunn testing (Benjamini/Hochberg correction) for continuous variables. For categorical variables, we applied the $\chi^2$ test followed by post-hoc pairwise chi-squared comparisons.

Statistical analyses were performed using the SciPy (version 1.9.0), scikit-posthocs (version 0.11.2), pengouin (version 0.5.5) and lifelines (version 0.27.8) libraries in Python, and P less than 0.05 was considered statistically significant.

## ACKNOWLEDGEMENTS

The FORCE registry is funded through grants from Additional Ventures and the Evan's Heart Fund.

## AUTHOR CONTRIBUTIONS

Guarantors of integrity of entire study, **T.Y., N.S.C., R.H.R., V.M.**; study concepts/study design or data acquisition or data analysis/interpretation, all authors; manuscript drafting or manuscript revision for important intellectual content, all authors; approval of final version of submitted manuscript, all authors; agrees to ensure any questions related to the work are appropriately resolved, all authors; literature research, **T.Y., R.K., T.C.S., V.M.**; clinical studies, **N.S.C., A.L.D., S.M., M.Q., M.A.F., R.K., C.Z.L., J.D.R., T.C.S., J.W., R.H.R.**; experimental studies, **T.Y., N.S.C., M.G, G.F.M., J.D.R., J.A.S., R.H.R.**; statistical analysis, **T.Y., G.F.M., V.M.**; and manuscript editing, **T.Y., N.S.C., A.L.D., M.A.F., R.K., C.Z.L., J.D.R., T.C.S., J.A.S., R.H.R., V.M.**

**T.Y.** and **N.S.C.** contributed equally to this work.

# FIGURES

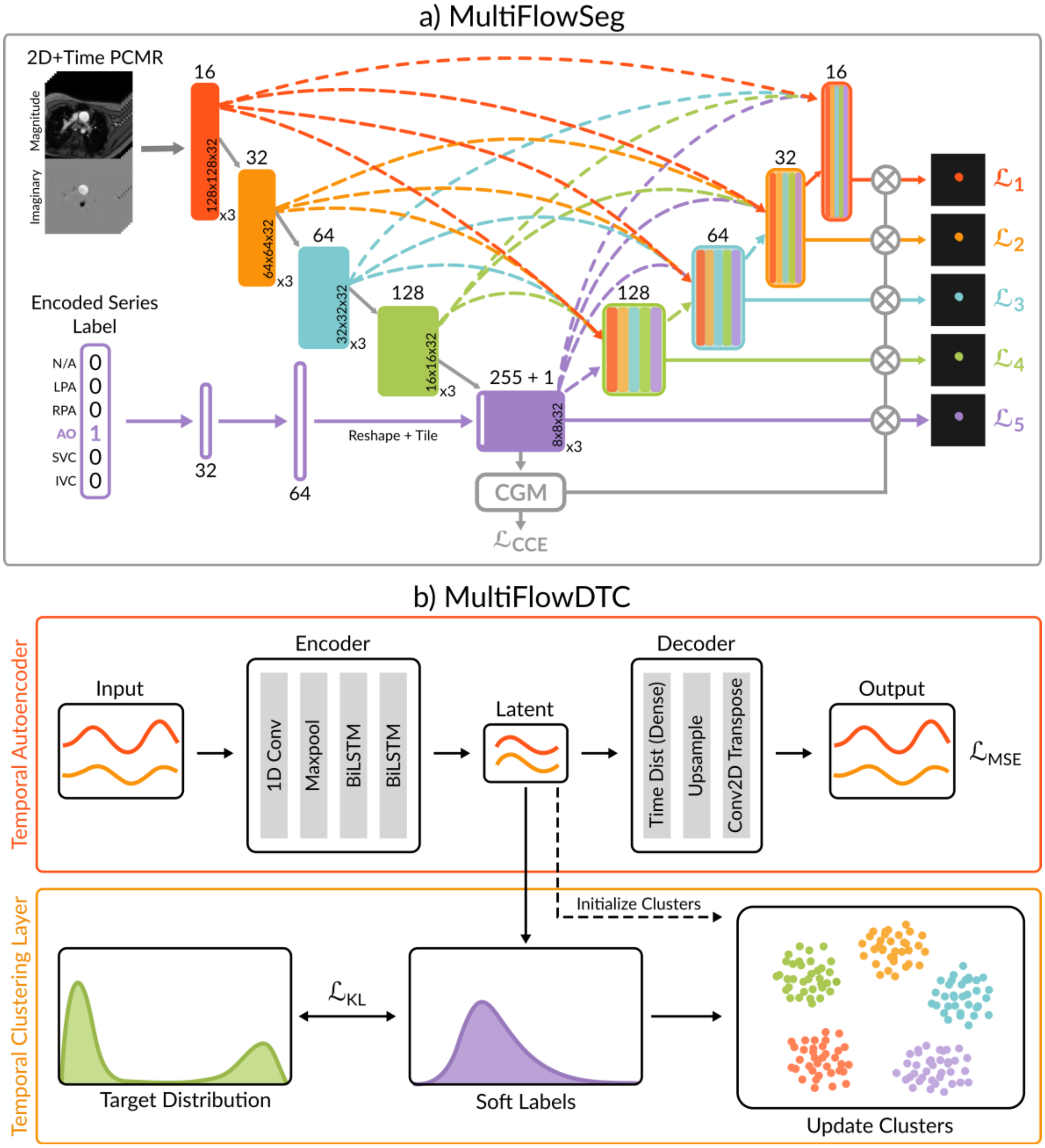

***Figure 1. a) Illustration of MultiFlowSeg – 2D+time Tunable UNet3+ with multi-class classification guidance. b) MultFlowDTC - Deep Temporal Clustering Model Architecture***

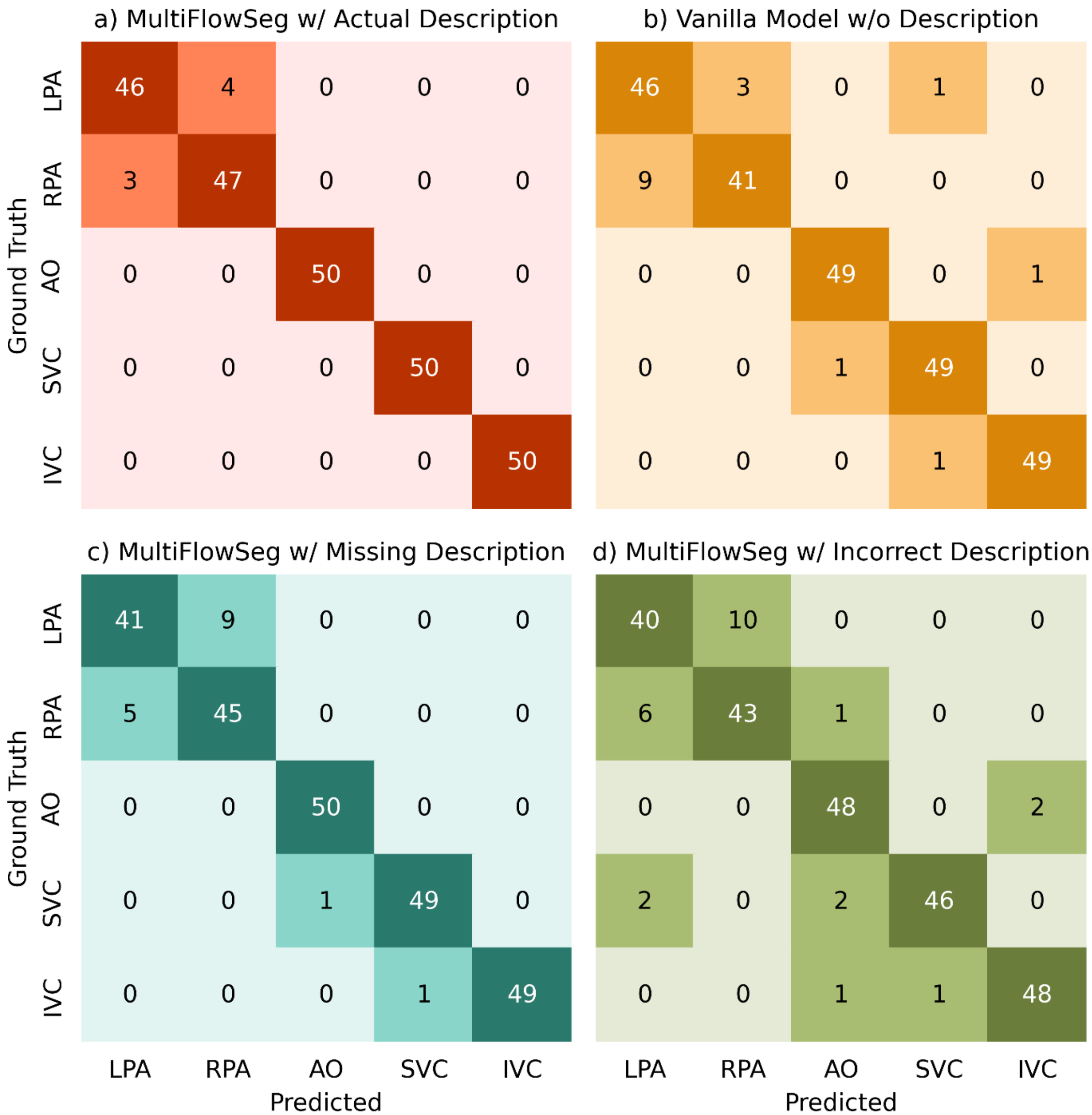

| a) Ground Truth \ Predicted | LPA | RPA | AO | SVC | IVC |
|---|---|---|---|---|---|
| LPA | 46 | 4 | 0 | 0 | 0 |
| RPA | 3 | 47 | 0 | 0 | 0 |
| AO | 0 | 0 | 50 | 0 | 0 |
| SVC | 0 | 0 | 0 | 50 | 0 |
| IVC | 0 | 0 | 0 | 0 | 50 |

| b) Ground Truth \ Predicted | LPA | RPA | AO | SVC | IVC |
|---|---|---|---|---|---|
| LPA | 46 | 3 | 0 | 1 | 0 |
| RPA | 9 | 41 | 0 | 0 | 0 |
| AO | 0 | 0 | 49 | 0 | 1 |
| SVC | 0 | 0 | 1 | 49 | 0 |
| IVC | 0 | 0 | 0 | 1 | 49 |

| c) Ground Truth \ Predicted | LPA | RPA | AO | SVC | IVC |
|---|---|---|---|---|---|
| LPA | 41 | 9 | 0 | 0 | 0 |
| RPA | 5 | 45 | 0 | 0 | 0 |
| AO | 0 | 0 | 50 | 0 | 0 |
| SVC | 0 | 0 | 1 | 49 | 0 |
| IVC | 0 | 0 | 0 | 1 | 49 |

| d) Ground Truth \ Predicted | LPA | RPA | AO | SVC | IVC |
|---|---|---|---|---|---|
| LPA | 40 | 10 | 0 | 0 | 0 |
| RPA | 6 | 43 | 1 | 0 | 0 |
| AO | 0 | 0 | 48 | 0 | 2 |
| SVC | 2 | 0 | 2 | 46 | 0 |
| IVC | 0 | 0 | 1 | 1 | 48 |

***Figure 2. Confusion matrix for showing the classification accuracy of our model on the unseen 50 test set. LPA – Left Pulmonary Artery, RPA – Right Pulmonary Artery, AO – Aorta, SVC – Superior Vena Cava, IVC – Inferior Vena Cava.***

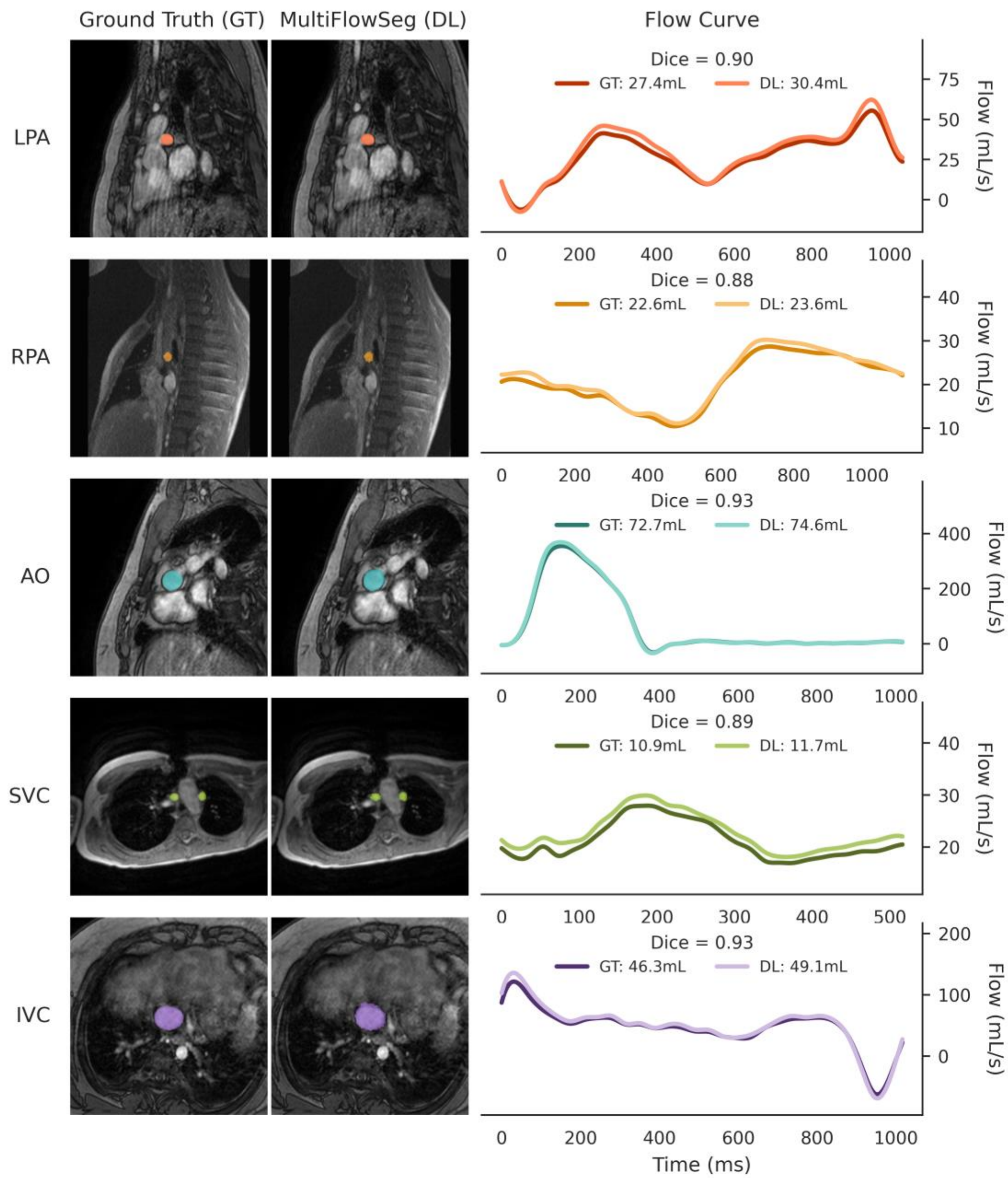

***Figure 3. Comparison of ground truth and predicted segmentations for each vessel. The median Dice score example is shown in the first frame, with flow curves for both ground truth and predictions compared. LPA – Left Pulmonary Artery, RPA – Right Pulmonary Artery, AO – Aorta, SVC – Superior Vena Cava, IVC – Inferior Vena Cava.***

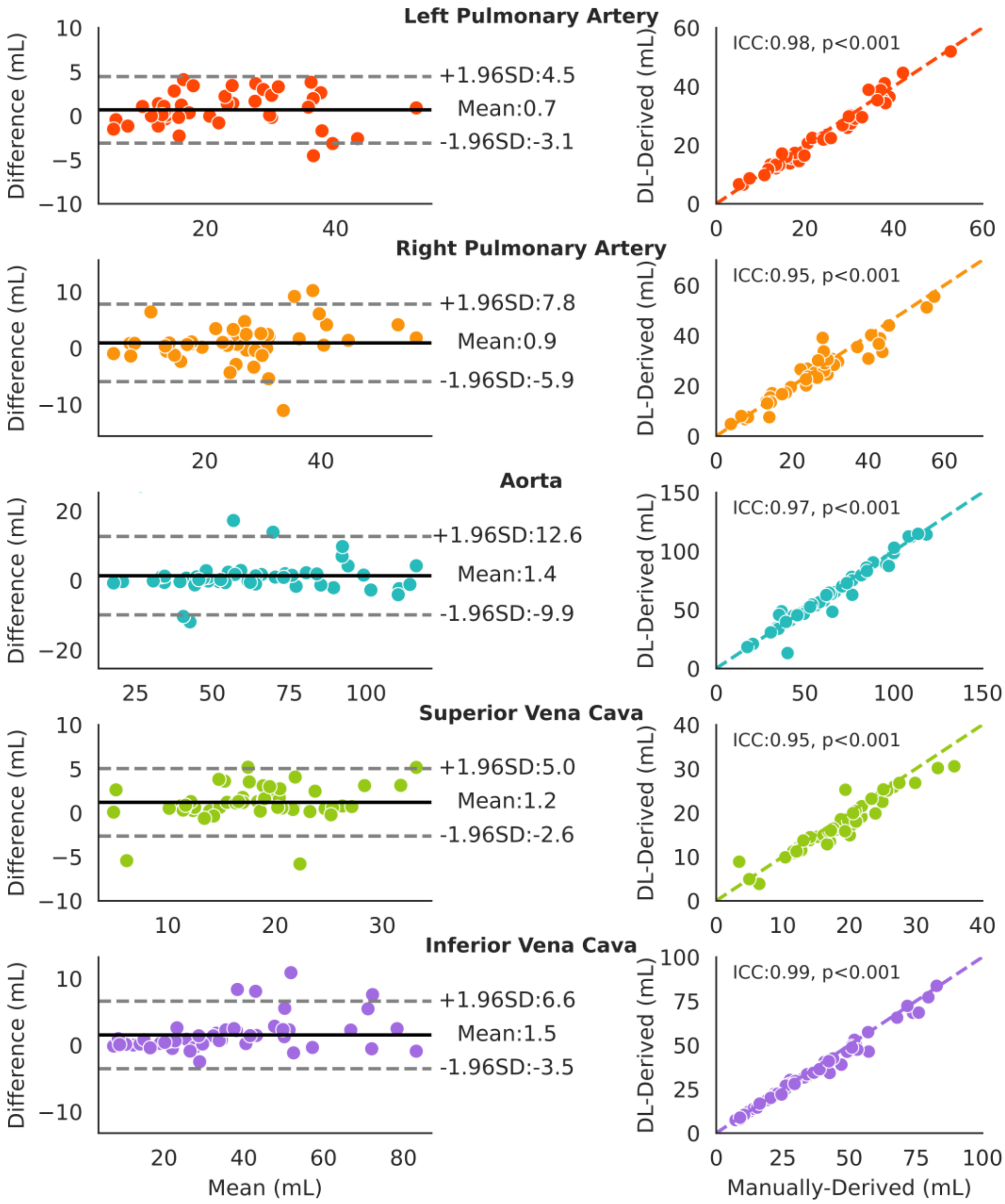

***Figure 4. Bland-Altman and correlation plots comparing flow derived from deep learning (DL) and manual ground truth segmentations across 50 test sets per vessel.***

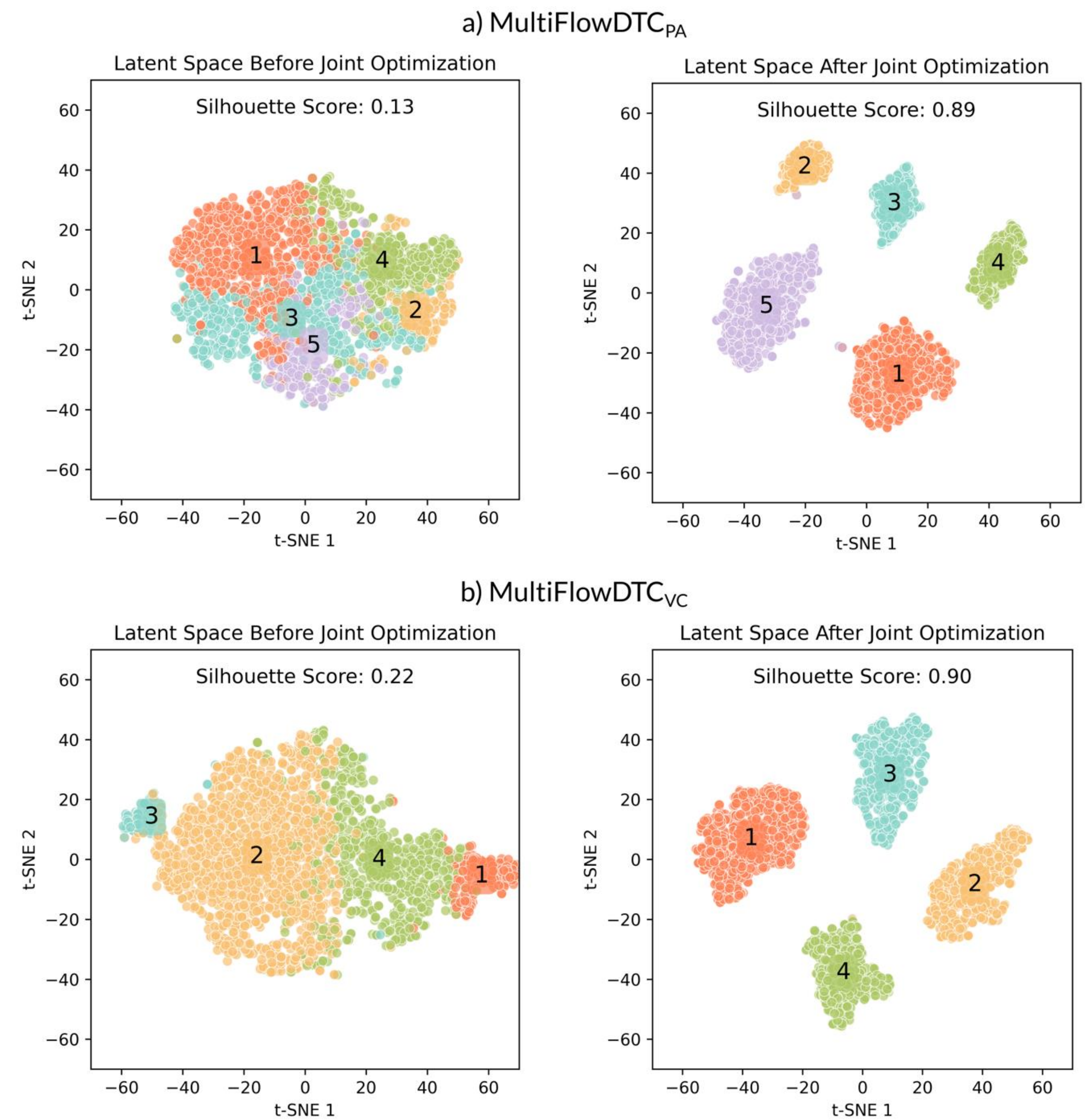

***Figure 5. t-distributed stochastic neighbor embedding (t-SNE) visualization of latent space clusters from the pulmonary artery and vena cava MultiFlowDTC models before and after joint optimization of the temporal autoencoder and clustering layer.***

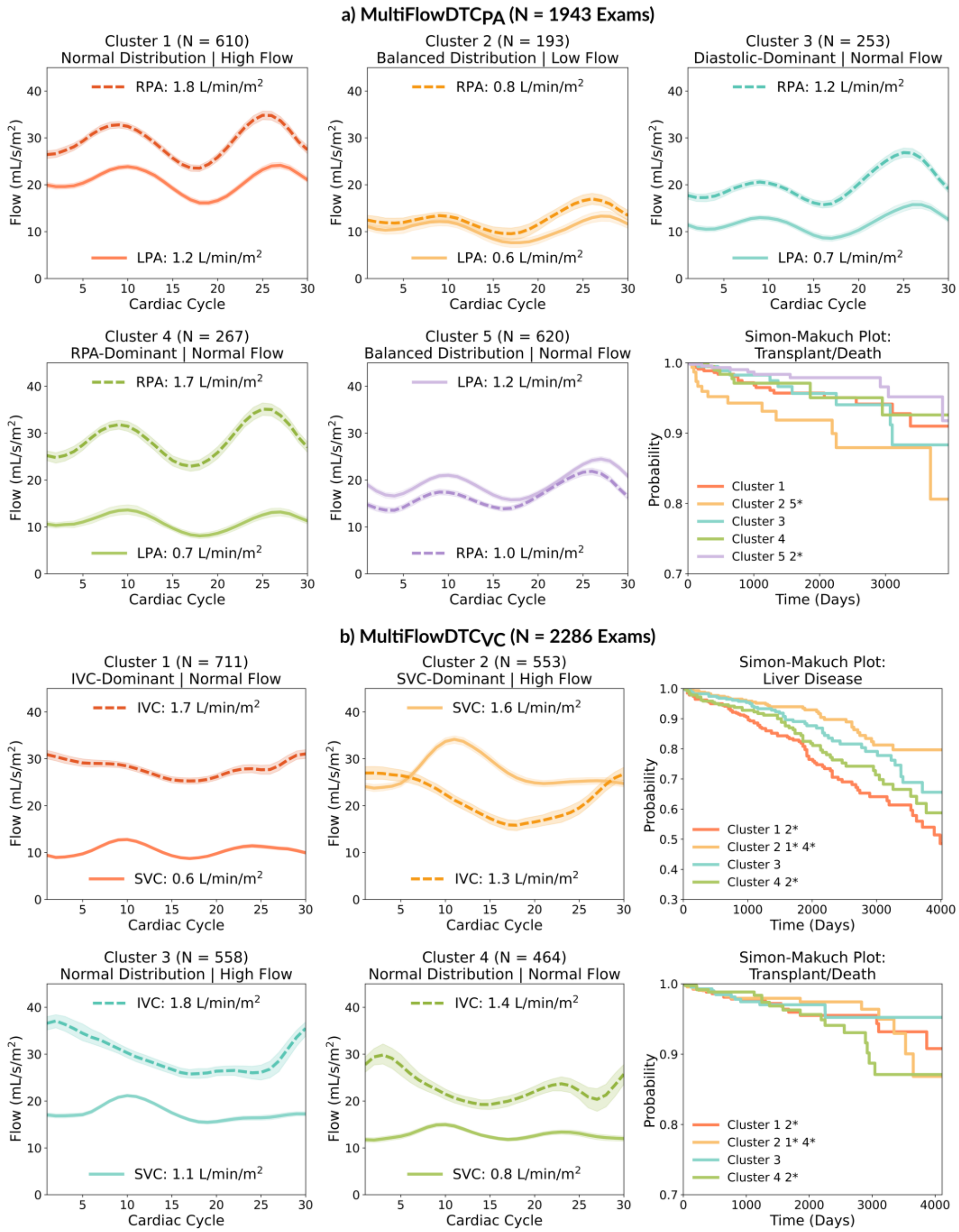

***Figure 6. Flow phenotypes from pulmonary artery and vena cava MultiFlowDTC models. Centroid flow curves and mean vessel flow are shown per cluster. Simon-Makuch plots illustrate survival probability for each cluster for liver disease and transplant/death (liver disease not shown for the pulmonary artery model due to non-significant differences). LPA – Left Pulmonary Artery, RPA – Right Pulmonary Artery, SVC – Superior Vena Cava, IVC – Inferior Vena Cava. * indicate significant differences between clusters according to time-varying Cox Regression.***

# APPENDIX

***Supplementary Table 1: Pipeline segmentation accuracy differences across variables***

| Vessel | Age | Bilateral Aorta | Bilateral SVC | Situs Type | Diagnosis | Heterotaxy | Image Quality |
|---|---|---|---|---|---|---|---|
| **All** | **<0.001** | **<0.001** | **<0.001** | **<0.001** | **<0.001** | **<0.001** | **<0.001** |
| **LPA** | 0.012 | **0.015** | **0.004** | **<0.001** | **<0.001** | **<0.001** | 0.11 |
| **RPA** | 0.312 | 0.047 | **<0.001** | **<0.001** | **<0.001** | **<0.001** | **<0.001** |
| **Aorta** | **<0.001** | **<0.001** | 0.227 | 0.312 | 0.029 | 0.76 | 0.829 |
| **SVC** | 0.97 | **0.002** | **<0.001** | **<0.001** | 0.097 | **<0.001** | **0.027** |
| **IVC** | **<0.001** | 0.249 | 0.173 | 0.892 | 0.322 | **<0.001** | 0.101 |

***Supplementary Table 2: Demographics of the 5 clusters for the pulmonary artery MultiFlowDTC model***

| | **Cluster 1 (n = 610)** | **Cluster 2 (n = 193)** | **Cluster 3 (n = 253)** | **Cluster 4 (n = 267)** | **Cluster 5 (n = 620)** | **p** |
|---|---|---|---|---|---|---|
| **BSA ($m^2$)** | 1.4 (1.1–1.7) | 1.6 (1.2–1.8) | 1.6 (1.3–1.8) | 1.5 (1.0–1.8) | 1.5 (1.2–1.8) | **<0.001** |
| **Age** | 13.7 (10.4–18.4) | 17.0 (11.4–23.3) | 16.6 (12.6–22.5) | 14.5 (10.8–19.3) | 16.1 (12.0–22.8) | **<0.001** |
| **Adult** | 160 (26) | 88 (46) | 102 (40) | 81 (30) | 248 (40) | |
| **Child (<18)** | 450 (74) | 105 (54) | 151 (60) | 186 (70) | 372 (60) | |
| **Male** | 395 (65) | 102 (53) | 152 (60) | 174 (65) | 352 (57) | **0.004** |
| **Diagnosis** | | | | | | **<0.001** |
| **HLHS** | 234 (38) | 74 (38) | 99 (39) | 117 (44) | 179 (29) | |
| **Tricuspid Atresia** | 94 (15) | 29 (15) | 43 (17) | 39 (15) | 141 (23) | |
| **Double Outlet Right Ventricle** | 65 (11) | 26 (13) | 37 (15) | 22 (8) | 58 (9) | |
| **Double Inlet Left Ventricle** | 94 (15) | 11 (6) | 25 (10) | 33 (12) | 71 (11) | |
| **Other** | 123 (20) | 53 (27) | 49 (19) | 56 (21) | 171 (28) | |
| **Fontan Type** | | | | | | **<0.001** |
| **Lateral** | 280 (46) | 79 (41) | 128 (51) | 141 (53) | 237 (38) | |
| **Extracardiac** | 283 (46) | 92 (48) | 109 (43) | 115 (43) | 272 (44) | |
| **Other/Unknown** | 47 (8) | 22 (11) | 16 (6) | 11 (4) | 111 (18) | |
| **Systemic Ventricular Circulation** | | | | | | 0.147 |
| **Both** | 300 (49) | 106 (55) | 123 (49) | 127 (48) | 300 (48) | |
| **Right** | 199 (33) | 61 (32) | 86 (34) | 92 (34) | 174 (28) | |
| **Left** | 110 (18) | 26 (13) | 44 (17) | 47 (18) | 145 (23) | |
| **Unknown** | 1 (0) | 0 (0) | 0 (0) | 1 (0) | 1 (0) | |
| **Heterotaxy** | 36 (6) | 23 (12) | 17 (7) | 8 (3) | 67 (11) | **0.01** |
| **Situs Type** | | | | | | **0.002** |
| **Levo** | 581 (95) | 182 (94) | 245 (97) | 261 (98) | 566 (91) | |
| **Dextro** | 22 (4) | 9 (5) | 6 (2) | 5 (2) | 48 (8) | |
| **Meso** | 7 (1) | 2 (1) | 2 (1) | 1 (0) | 6 (1) | |
| **Heart Rate (BPM)** | 82 (71–92) | 74 (66–88) | 70 (63–85) | 78 (68–91) | 74 (64–84) | **<0.001** |
| **SpO2 (%)** | 95 (93–97) (N = 58) | 91 (89–94) (N = 13) | 94 (90–96) (N = 18) | 95 (93–98) (N = 17) | 94 (92–97) (N = 48) | **0.011** |
| **Collateral (%)** | 18 (10–28) (N = 463) | 25 (17–39) (N = 161) | 23 (14–34) (N = 211) | 23 (13–36) (N = 216) | 19 (11–32) (N = 472) | **<0.001** |
| **EDVi (mL/$m^2$)** | 101.5 (84.2–124.0) (N = 595) | 105.6 (82.3–135.5) (N = 192) | 102.8 (82.3–125.9) | 102.5 (84.8–130.8) (N = 265) | 94.5 (76.1–115.4) (N = 610) | **<0.001** |
| **ESVi (mL/$m^2$)** | 47.5 (36.5–62.2) (N = 594) | 54.0 (39.5–73.0) (N = 191) | 51.3 (37.5–66.8) (N = 251) | 49.8 (35.9–67.2) (N = 265) | 44.4 (32.5–58.7) (N = 610) | **<0.001** |
| **EF (%)** | 53 (47–58) (N = 594) | 49 (42–53) (N = 191) | 51 (45–56) (N = 251) | 52 (44–59) (N = 265) | 53 (45–58) (N = 610) | **<0.001** |
| **EDP (mmHg)** | 8.0 (6.0–10.0) (N = 56) | 9.0 (7.0–11.0) (N = 45) | 8.0 (6.0–10.0) (N = 43) | 8.5 (6.0–12.0) (N = 36) | 9.0 (7.0–11.0) (N = 84) | 0.801 |
| **VO2 PP (%)** | 67 (59–76) (N = 97) | 58 (51–70) (N = 30) | 59 (52–70) (N = 43) | 61 (52–72) (N = 41) | 66 (56–78) (N = 124) | **0.016** |
| **LPA/(LPA + RPA)** | 0.42 (0.37–0.45) | 0.45 (0.41–0.49) | 0.38 (0.33–0.41) | 0.3 (0.23–0.35) | 0.52 (0.48–0.58) | **<0.001** |
| **Indexed AO Flow Rate (L/min/m$^2$)** | 3.5 (3.0–4.0) | 2.5 (2.1–3.1) | 2.8 (2.4–3.3) | 3.2 (2.7–3.7) | 2.9 (2.5–3.3) | **<0.001** |
| **Indexed LPA Flow Rate (L/min/m$^2$)** | 1.2 (1.0–1.4) | 0.7 (0.5–0.8) | 0.7 (0.6–0.9) | 0.7 (0.5–0.8) | 1.1 (1.0–1.3) | **<0.001** |
| **Indexed RPA Flow Rate (L/min/m$^2$)** | 1.7 (1.5–1.9) | 0.8 (0.7–0.9) | 1.2 (1.1–1.3) | 1.7 (1.5–1.9) | 1.0 (0.9–1.2) | **<0.001** |
| **Indexed SVC Flow Rate (L/min/m$^2$)** | 1.1 (0.9–1.3) (N = 592) | 0.7 (0.6–0.9) (N = 186) | 0.8 (0.6–1.0) (N = 246) | 0.9 (0.7–1.2) (N = 259) | 0.8 (0.7–1.1) (N = 604) | **<0.001** |
| **Indexed IVC Flow Rate (L/min/m$^2$)** | 1.9 (1.5–2.2) (N = 604) | 1.1 (0.9–1.4) (N = 186) | 1.4 (1.1–1.7) (N = 249) | 1.5 (1.1–1.8) (N = 263) | 1.5 (1.2–1.8) (N = 605) | **<0.001** |
| **VC Clusters** | | | | | | |
| **Cluster 1** | 110 (18) | 68 (35) | 111 (44) | 86 (32) | 207 (33) | |
| **Cluster 2** | 176 (29) | 28 (15) | 35 (14) | 69 (26) | 107 (17) | |
| **Cluster 3** | 219 (36) | 19 (10) | 31 (12) | 63 (24) | 137 (22) | |
| **Cluster 4** | 81 (13) | 64 (33) | 65 (26) | 37 (14) | 138 (22) | |
| **N/A** | 24 (4) | 14 (7) | 11 (4) | 12 (4) | 31 (5) | |

***Supplementary Table 3: Statistical Differences in Demographics of the 5 clusters for the PA model***

| Variable | Overall | 1 vs 2 | 1 vs 3 | 1 vs 4 | 1 vs 5 | 2 vs 3 | 4 vs 2 | 4 vs 3 | 4 vs 5 | 5 vs 2 | 5 vs 3 |
|---|---|---|---|---|---|---|---|---|---|---|---|
| **Male** | **0.004** | **0.004** | 0.222 | 0.967 | **0.005** | 0.152 | **0.01** | 0.268 | **0.024** | 0.381 | 0.411 |
| **Diagnosis** | **<0.001** | **0.012** | **0.041** | 0.393 | **<0.001** | 0.383 | **0.022** | 0.145 | **<0.001** | **0.003** | **0.001** |
| **Fontan Type** | **<0.001** | 0.223 | 0.218 | 0.125 | **<0.001** | 0.198 | **0.017** | 0.284 | **<0.001** | 0.622 | **0.001** |
| **Systemic Ventricular Circulation** | 0.147 | 0.373 | 0.903 | 0.881 | 0.094 | 0.348 | 0.334 | 0.804 | 0.12 | **0.028** | 0.138 |
| **Heterotaxy** | **0.01** | **0.031** | 0.186 | 0.41 | **0.009** | 0.186 | **0.006** | 0.164 | **0.004** | 0.993 | 0.186 |
| **Situs Type** | **0.002** | 0.797 | 0.575 | 0.207 | **0.007** | 0.395 | 0.153 | 0.758 | **0.002** | 0.343 | **0.011** |
| **BSA** | **<0.001** | **0.017** | **0.001** | 0.696 | **<0.001** | 0.696 | 0.108 | **0.017** | **0.017** | 0.79 | 0.718 |
| **Age** | **<0.001** | **<0.001** | **<0.001** | 0.8 | **<0.001** | 0.968 | **0.002** | **0.001** | **<0.001** | 0.968 | 0.968 |
| **Heart Rate** | **<0.001** | **<0.001** | **<0.001** | 0.05 | **<0.001** | 0.087 | 0.05 | **<0.001** | **<0.001** | 0.246 | 0.33 |
| **SpO2** | **0.011** | 0.998 | 0.998 | 0.998 | 0.998 | 0.998 | 0.998 | 0.998 | 0.998 | 0.998 | 0.998 |
| **Collateral** | **<0.001** | **<0.001** | **0.011** | **0.004** | 0.122 | 0.158 | 0.251 | 0.673 | 0.099 | **0.005** | 0.182 |
| **EDVi** | **<0.001** | 0.441 | 0.935 | 0.556 | **<0.001** | 0.441 | 0.751 | 0.556 | **<0.001** | **<0.001** | **0.002** |
| **ESVi** | **<0.001** | **0.002** | 0.168 | 0.351 | **0.006** | 0.13 | 0.057 | 0.661 | **0.002** | **<0.001** | **0.001** |
| **EF** | **<0.001** | **<0.001** | **<0.001** | 0.176 | 0.424 | **0.033** | **<0.001** | **0.043** | 0.424 | **<0.001** | **0.002** |
| **EDP** | 0.801 | 0.993 | 0.993 | 0.993 | 0.993 | 0.993 | 0.993 | 0.993 | 0.993 | 0.993 | 0.993 |
| **VO2 PP** | **0.016** | 0.956 | 0.956 | 0.956 | 0.956 | 0.956 | 0.956 | 0.956 | 0.956 | 0.956 | 0.956 |
| **LPA/(LPA + RPA)** | **<0.001** | **<0.001** | **<0.001** | **<0.001** | **<0.001** | **<0.001** | **<0.001** | **<0.001** | **<0.001** | **<0.001** | **<0.001** |
| **Indexed AO Flow Rate** | **<0.001** | **<0.001** | **<0.001** | **<0.001** | **<0.001** | **0.001** | **<0.001** | **<0.001** | **<0.001** | **<0.001** | 0.099 |
| **Indexed SVC Flow Rate** | **<0.001** | **<0.001** | **<0.001** | **<0.001** | **<0.001** | **0.026** | **<0.001** | **<0.001** | **0.001** | **<0.001** | **0.004** |
| **Indexed IVC Flow Rate** | **<0.001** | **<0.001** | **<0.001** | **<0.001** | **<0.001** | **<0.001** | **<0.001** | 0.056 | 0.355 | **<0.001** | **0.002** |
| **Indexed LPA Flow Rate** | **<0.001** | **<0.001** | **<0.001** | **<0.001** | **0.001** | **0.022** | 0.178 | 0.282 | **<0.001** | **<0.001** | **<0.001** |
| **Indexed RPA Flow Rate** | **<0.001** | **<0.001** | **<0.001** | 0.34 | **<0.001** | **<0.001** | **<0.001** | **<0.001** | **<0.001** | **<0.001** | **<0.001** |

***Supplementary Table 4: Demographics of the 4 clusters for the VC model***

| | Cluster 1 (n = 711) | Cluster 2 (n = 553) | Cluster 3 (n = 558) | Cluster 4 (n = 464) | p |
|---|---|---|---|---|---|
| **BSA (m²)** | 1.8 (1.6–2.0) | 0.9 (0.7–1.1) | 1.4 (1.2–1.6) | 1.6 (1.4–1.8) | **<0.001** |
| **Age** | 20.6 (16.7–26.5) | 8.2 (5.4–10.8) | 13.5 (11.3–16.4) | 17.2 (14.3–22.5) | **<0.001** |
| **Adult** | 467 (66) | 14 (3) | 92 (16) | 205 (44) | |
| **Child (<18)** | 244 (34) | 539 (97) | 466 (84) | 259 (56) | |
| **Male** | 417 (59) | 347 (63) | 355 (64) | 265 (57) | 0.082 |
| **Diagnosis** | | | | | **0.006** |
| **HLHS** | 234 (33) | 232 (42) | 206 (37) | 158 (34) | |
| **Tricuspid Atresia** | 127 (18) | 85 (15) | 90 (16) | 78 (17) | |
| **Double Outlet Right Ventricle** | 72 (10) | 53 (10) | 68 (12) | 55 (12) | |
| **Double Inlet Left Ventricle** | 93 (13) | 55 (10) | 63 (11) | 44 (9) | |
| **Other** | 185 (26) | 128 (23) | 131 (23) | 129 (28) | |
| **Fontan Type** | | | | | **<0.001** |
| **Lateral** | 401 (56) | 147 (27) | 242 (43) | 205 (44) | |
| **Extracardiac** | 226 (32) | 358 (65) | 274 (49) | 198 (43) | |
| **Other/Unknown** | 84 (12) | 48 (9) | 42 (8) | 61 (13) | |
| **Systemic Ventricular Circulation** | | | | | **0.021** |
| **Both** | 379 (53) | 257 (46) | 273 (49) | 222 (48) | |
| **Right** | 197 (28) | 202 (37) | 172 (31) | 158 (34) | |
| **Left** | 135 (19) | 94 (17) | 111 (20) | 84 (18) | |
| **Unknown** | 0 (0) | 0 (0) | 2 (0) | 0 (0) | |
| **Heterotaxy** | 74 (10) | 65 (12) | 41 (7) | 63 (14) | **0.004** |
| **Situs Type** | | | | | 0.56 |
| **Levo** | 640 (90) | 508 (92) | 517 (93) | 419 (90) | |
| **Dextro** | 64 (9) | 41 (7) | 37 (7) | 38 (8) | |
| **Meso** | 7 (1) | 4 (1) | 4 (1) | 7 (2) | |
| **Heart Rate (BPM)** | 71 (63–80) | 90 (79–105) | 80 (71–90) | 71 (62–81) | **<0.001** |
| **SpO2 (%)** | 94 (90–95) (N = 56) | 97 (94–99) (N = 36) | 95 (93–97) (N = 52) | 94 (92–96) (N = 36) | **0.001** |
| **Collateral (%)** | 20 (12–30) (N = 593) | 24 (14–36) (N = 468) | 18 (9–29) (N = 430) | 24 (13–38) (N = 396) | **<0.001** |
| **EDVi (mL/m2)** | 98.8 (79.7–123.2) (N = 708) | 98.6 (79.9–123.6) (N = 548) | 101.4 (81.9–122.0) (N = 544) | 99.7 (83.4–127.6) (N = 462) | 0.671 |
| **ESVi (mL/m²)** | 47.5 (34.8–64.9) (N = 708) | 46.7 (35.7–62.1) (N = 545) | 47.3 (36.3–62.2) (N = 542) | 50.2 (36.7–68.0) (N = 462) | 0.138 |
| **EF (%)** | 51 (45–57) (N = 708) | 53 (46–58) (N = 545) | 52 (45–58) (N = 542) | 51 (43–56) (N = 462) | **0.002** |
| **EDP (mmHg)** | 10.0 (8.0–12.0) (N = 89) | 8.0 (5.0–10.0) (N = 111) | 9.0 (7.0–11.0) (N = 56) | 9.0 (6.0–11.0) (N = 65) | **<0.001** |
| **VO2 PP (%)** | 60 (53–72) (N = 132) | 72 (66–86) (N = 42) | 68 (58–81) (N = 104) | 63 (54–71) (N = 93) | **<0.001** |
| **SVC/(SVC + IVC)** | 0.28 (0.24–0.31) | 0.56 (0.48–0.65) | 0.37 (0.33–0.43) | 0.37 (0.32–0.41) | **<0.001** |
| **Indexed AO Flow Rate (L/min/m²)** | 2.8 (2.4–3.2) | 3.7 (3.1–4.3) | 3.3 (2.9–3.7) | 2.7 (2.3–3.2) | **<0.001** |
| **Indexed LPA Flow Rate (L/min/m²)** | 0.9 (0.7–1.1) (N = 667) | 1.0 (0.8–1.3) (N = 492) | 1.1 (0.9–1.3) (N = 521) | 0.9 (0.7–1.1) (N = 438) | **<0.001** |
| **Indexed RPA Flow Rate (L/min/m²)** | 1.2 (0.9–1.5) (N = 599) | 1.4 (1.1–1.8) (N = 445) | 1.5 (1.2–1.8) (N = 493) | 1.2 (0.9–1.4) (N = 397) | **<0.001** |
| **Indexed SVC Flow Rate (L/min/m²)** | 0.7 (0.5–0.7) | 1.5 (1.3–1.9) | 1.1 (1.0–1.1) | 0.8 (0.7–0.9) | **<0.001** |
| **Indexed IVC Flow Rate (L/min/m²)** | 1.7 (1.4–1.9) | 1.2 (0.9–1.6) | 1.8 (1.4–2.1) | 1.4 (1.0–1.7) | **<0.001** |
| **PA Clusters** | | | | | |
| **Cluster 1** | 110 (15) | 176 (32) | 219 (39) | 81 (17) | |
| **Cluster 2** | 68 (10) | 28 (5) | 19 (3) | 64 (14) | |
| **Cluster 3** | 111 (16) | 35 (6) | 31 (6) | 65 (14) | |
| **Cluster 4** | 86 (12) | 69 (12) | 63 (11) | 37 (8) | |
| **Cluster 5** | 207 (29) | 107 (19) | 137 (25) | 138 (30) | |
| **N/A** | 129 (18) | 138 (25) | 89 (16) | 79 (17) | |

***Supplementary Table 5: Statistical Differences in Demographics of the 4 clusters for the VC DTC Model***

| Variable | Overall | 1 vs 3 | 2 vs 1 | 2 vs 3 | 2 vs 4 | 4 vs 1 | 4 vs 3 |
|---|---|---|---|---|---|---|---|
| **Male** | 0.082 | 0.081 | 0.155 | 0.811 | 0.078 | 0.644 | **0.04** |
| **Diagnosis** | **0.006** | **0.023** | 0.055 | 0.25 | **0.047** | 0.272 | **0.013** |
| **Fontan Type** | **<0.001** | **<0.001** | **<0.001** | **<0.001** | **<0.001** | **<0.001** | **<0.001** |
| **Systemic Ventricular Circulation** | **0.021** | 0.18 | **0.004** | 0.099 | 0.7 | 0.064 | 0.397 |
| **Heterotaxy** | **0.004** | 0.166 | **0.016** | **0.024** | **0.042** | 0.569 | **0.012** |
| **Situs Type** | 0.56 | 0.259 | 0.521 | 0.877 | 0.427 | 0.65 | 0.291 |
| **BSA** | **<0.001** | **<0.001** | **<0.001** | **<0.001** | **<0.001** | **<0.001** | **<0.001** |
| **Age** | **<0.001** | **<0.001** | **<0.001** | **<0.001** | **<0.001** | **<0.001** | **<0.001** |
| **Heart Rate** | **<0.001** | **<0.001** | **<0.001** | **<0.001** | **<0.001** | 0.777 | **<0.001** |
| **SpO2** | **0.001** | 0.944 | 0.944 | 0.944 | 0.944 | 0.944 | 0.944 |
| **Collateral** | **<0.001** | **0.03** | **0.004** | **<0.001** | 0.766 | **0.014** | **<0.001** |
| **EDVi** | 0.671 | 0.765 | 0.888 | 0.765 | 0.765 | 0.765 | 0.765 |
| **ESVi** | 0.138 | 0.989 | 0.641 | 0.641 | 0.156 | 0.207 | 0.207 |
| **EF** | **0.002** | 0.456 | 0.125 | 0.426 | **0.002** | 0.054 | **0.017** |
| **EDP** | **<0.001** | 0.973 | 0.973 | 0.973 | 0.973 | 0.973 | 0.982 |
| **VO2 PP** | **<0.001** | 0.901 | 0.901 | 0.901 | 0.901 | 0.944 | 0.901 |
| **SVC/(SVC + IVC)** | **<0.001** | **<0.001** | **<0.001** | **<0.001** | **<0.001** | **<0.001** | 0.168 |
| **Indexed AO Flow Rate** | **<0.001** | **<0.001** | **<0.001** | **<0.001** | **<0.001** | 0.283 | **<0.001** |
| **Indexed SVC Flow Rate** | **<0.001** | **<0.001** | **<0.001** | **<0.001** | **<0.001** | **<0.001** | **<0.001** |
| **Indexed IVC Flow Rate** | **<0.001** | **0.004** | **<0.001** | **<0.001** | **0.001** | **<0.001** | **<0.001** |
| **Indexed LPA Flow Rate** | **<0.001** | **<0.001** | **<0.001** | **0.003** | **<0.001** | 0.519 | **<0.001** |
| **Indexed RPA Flow Rate** | **<0.001** | **<0.001** | **<0.001** | 0.126 | **<0.001** | 0.346 | **<0.001** |

***Supplementary Table 6: Complete list of FORCE Investigator co-authors and affiliations.***

| Co-Author | Hospital |
| --- | --- |
| Aggarwal, M | Department of Pediatrics, St. Louis Children's Hospital, St. Louis, MO |
| Alsaied, T | The Heart and Vascular Institute, UPMC Children's Hospital of Pittsburgh, Pittsburgh, PA |
| Dorfman, AL | Congenital Heart Center, CS Mott Children's Hospital, Ann Arbor, MI |
| Doshi, A | Department of Pediatric and Congenital Cardiology, John Hopkins Hospital, Baltimore, MD |
| Files, MD | Division of Pediatric Cardiology, Seattle Children’s Hospital, Seattle, WA |
| Fogel, M | Division of Cardiology, The Children's Hospital of Philadelphia, Pennsylvania, PA |
| Hegde, S | Division of Pediatric Cardiology, Rady Children's Hospital, San Diego, CA |
| Hoyer, A | Division of Pediatric Cardiology, University of Arizona, Tucson, AZ |
| Johnson, T | Division of Pediatric Cardiology, Riley Hospital for Children, Indianapolis, IN |
| Krishnamurthy, R | Department of Radiology, Nationwide Children's Hospital, Columbus, OH |
| Lam, CZ | Department of Diagnostic and Interventional Radiology, The Hospital for Sick Children, Toronto, Canada |
| Loke, Y | Division of Cardiology, Children's National Hospital, Washington DC |
| Marsden, AL | Department of Bioengineering & Pediatrics, Stanford University, Palo Alto, CA |
| Muthurangu, V | UCL Centre for Cardiovascular Imaging, Institute of Cardiovascular Science, London, UK |
| Olivieri, LJ | The Heart and Vascular Institute, UPMC Children's Hospital of Pittsburgh, Pittsburgh, PA |
| Quail, M | UCL Centre for Cardiovascular Imaging, Institute of Cardiovascular Science, London, UK |
| Raimondi, F | Congenital Cardiology Unit, Ospedale Papa Giovanni XXIII, Bergamo, Italy |
| Ramachandran , P | Division of Pediatric Cardiology, University of Kentucky, Lexington, KY |
| Rathod, RH | Department of Cardiology, Boston Children’s Hospital, Boston, MA |
| Renella, P | Division of Pediatric Cardiology, CHOC Children’s Hospital, Orange, CA |
| Renno, MS | Division of Pediatric Cardiology, Arkansas Children's Hospital, Little Rock, AR |
| Robinson, JD | Division of Pediatric Cardiology, Ann & Robert H. Lurie Children's Hospital, Chicago, IL |
| Ruchira, G | Division of Pediatric Cardiology, Cedars-Sinai Guerin Children's Hospital, Los Angeles, CA |
| Shah, A | Division of Pediatric Cardiology, New York-Presbyterian Morgan Stanley Children’s Hospital, New York, NY |
| Slesnick, TC | Division of Pediatric Cardiology, Children's Healthcare of Atlanta, Atlanta, GA |
| Soslow, JH | Division of Pediatric Cardiology, Vanderbilt University Medical Center, Nashville, TN |
| Steele, J | Division of Pediatric Cardiology, Yale New Haven Children's Hospital, New Haven, CT |
| Stern, KW | Division of Pediatric Cardiology, Mount Sinai Kravis Children's Hospital, New York, NY |
| Thattaliyath, B | Division of Pediatric Cardiology, Stead Family Children's Hospital, Iowa City, IA |
| Vaikom House, A | Division of Pediatric Cardiology, Oklahoma Children's Hospital, Oklahoma City, OK |
| Weigand, J | Division of Pediatric Cardiology, Texas Children's Hospital, Houston, TX |

***Supplementary Table 7: Data Dictionaries***

| Vessel | Tunable Input Data Dictionary | Reference Data Dictionary |
|---|---|---|
| **LPA** | ['lpa', 'l pa', 'lt ', 'left'] | ['lpa'] |
| **RPA** | ['rpa', 'r pa', 'rt ', 'right'] | ['rpa'] |
| **Aorta** | ['aorta', 'ao', 'aor', 'asc', 'aso', 'neo', 'aa', 'aao', 'native', 'aov', 'av ', 'qs', 'stj', 'dks'] | ['asc','aa','aao','ao','neo','stj','dks'] |
| **SVC** | ['svc', 'rsvc', 'lsvc', 'sup', 'superior', 'glenn', 'vfc', 'bdg'] | ['svc'] |
| **IVC** | ['ivc', 'dao', 'inferior', 'fontan', 'font','inf '] | ['ivc'] |
| **Other** | ['', 'pv', 'avv', 'vein', 'lpv', 'rpv'] | - |

*Supplementary Table 7: Demographics of Patients Used for Training and Validation of MultiFlowSeg and for Testing the Automated Pipeline*

| | No. of Patients Exams in Manually Segmented Dataset (N = 260) | No. of Patients Exams in Pipeline Validated Dataset (N = 2902) |
|---|---|---|
| **BSA ($m^2$)** | 1.6 (1.3–1.9) (N = 228) | 1.5 (1.1–1.8) (N = 2704) |
| **Age** | 16.7 (13.1–23.8) (N = 228) | 15.1 (10.6–20.6) (N = 2712) |
| **Adult** | 100 (38) | 909 (31) |
| **Child (<18)** | 128 (49) | 1803 (62) |
| **Unknown** | 32 (12) | 190 (7) |
| **Sex** | | |
| **Male** | 134 (52) | 1637 (56) |
| **Female** | 94 (36) | 1075 (37) |
| **Unknown** | 32 (12) | 190 (7) |
| **Diagnosis** | | |
| **HLHS** | 73 (28) | 950 (33) |
| **Tricuspid Atresia** | 36 (14) | 451 (16) |
| **Double Outlet Right Ventricle** | 23 (9) | 310 (11) |
| **Double Inlet Left Ventricle** | 32 (12) | 296 (10) |
| **Other** | 96 (37) | 895 (31) |
| **Fontan Type** | | |
| **Lateral** | 97 (37) | 1216 (42) |
| **Extracardiac** | 108 (42) | 1206 (42) |
| **Other/Unknown** | 55 (21) | 480 (16) |
| **Systemic Ventricular Circulation** | | |
| **Both** | 131 (50) | 1399 (48) |
| **Right** | 59 (23) | 810 (28) |
| **Left** | 38 (15) | 499 (17) |
| **Unknown** | 32 (12) | 194 (7) |
| **Heterotaxy** | | |
| **No** | 194 (75) | 2410 (83) |
| **Yes** | 32 (12) | 294 (10) |
| **Unknown** | 34 (13) | 198 (7) |
| **Situs Type** | | |
| **Levo** | 235 (90) | 2675 (92) |
| **Dextro** | 25 (10) | 202 (7) |
| **Meso** | 0 (0) | 25 (1) |

***Supplementary Table 8. Number of patients who underwent heart transplantation, died, experienced either death or transplant (whichever occurred first), and had liver disease in the pulmonary artery and vena cava dataset used for the MultiFlowDTC models.***

| | Number of Patients in Dataset with Outcome | | | |
|---|---|---|---|---|
| | **Transplant** | **Death** | **Transplant/Death** | **Liver Disease** |
| **MultiFlowDTC$_{PA}$** | 22 | 41 | 62 | 204 |
| **MultiFlowDTC$_{VC}$** | 23 | 44 | 64 | 239 |

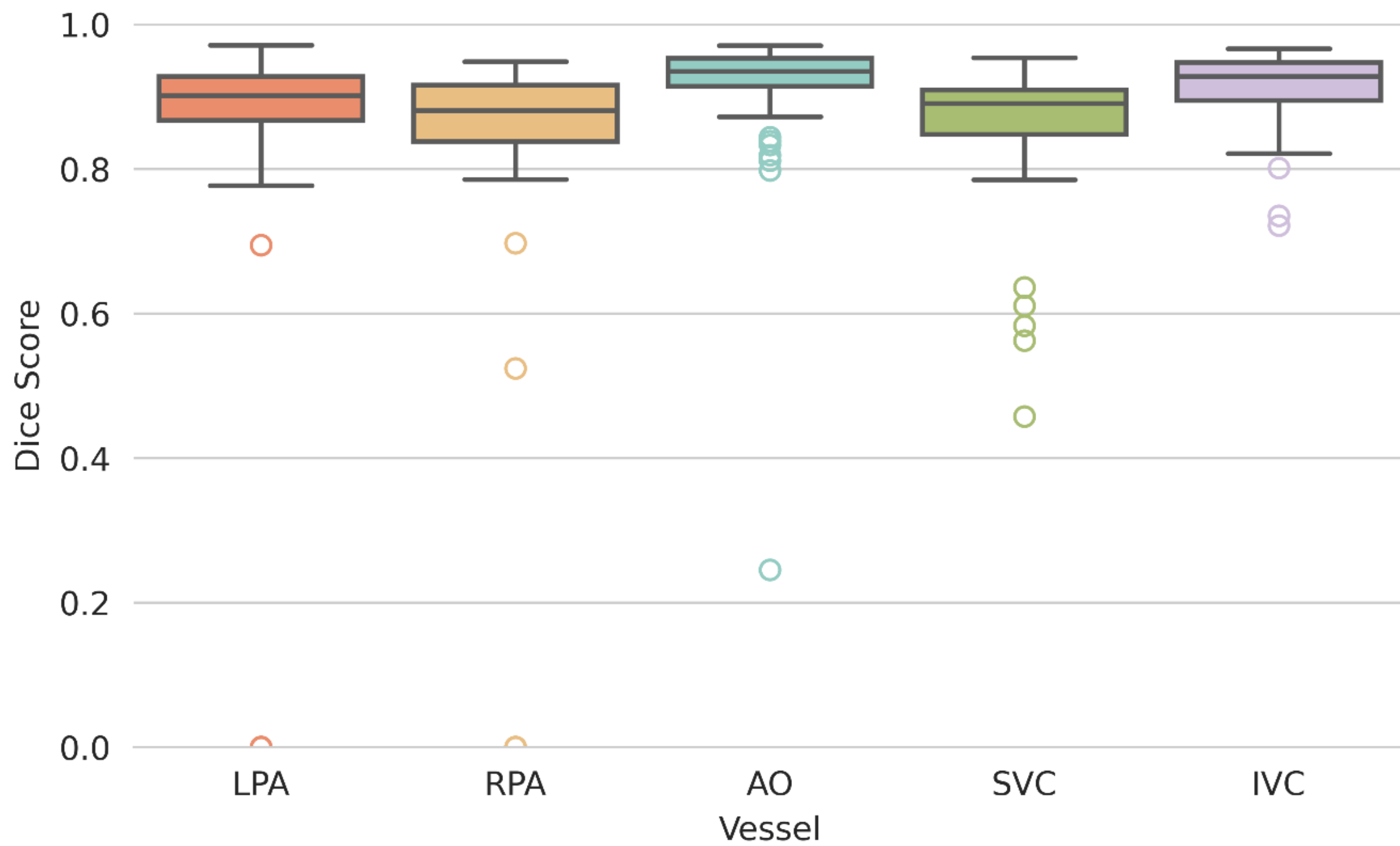

***Supplementary Figure 1: Box plot of the Dice scores for each vessel on the 50 ground truth test sets***

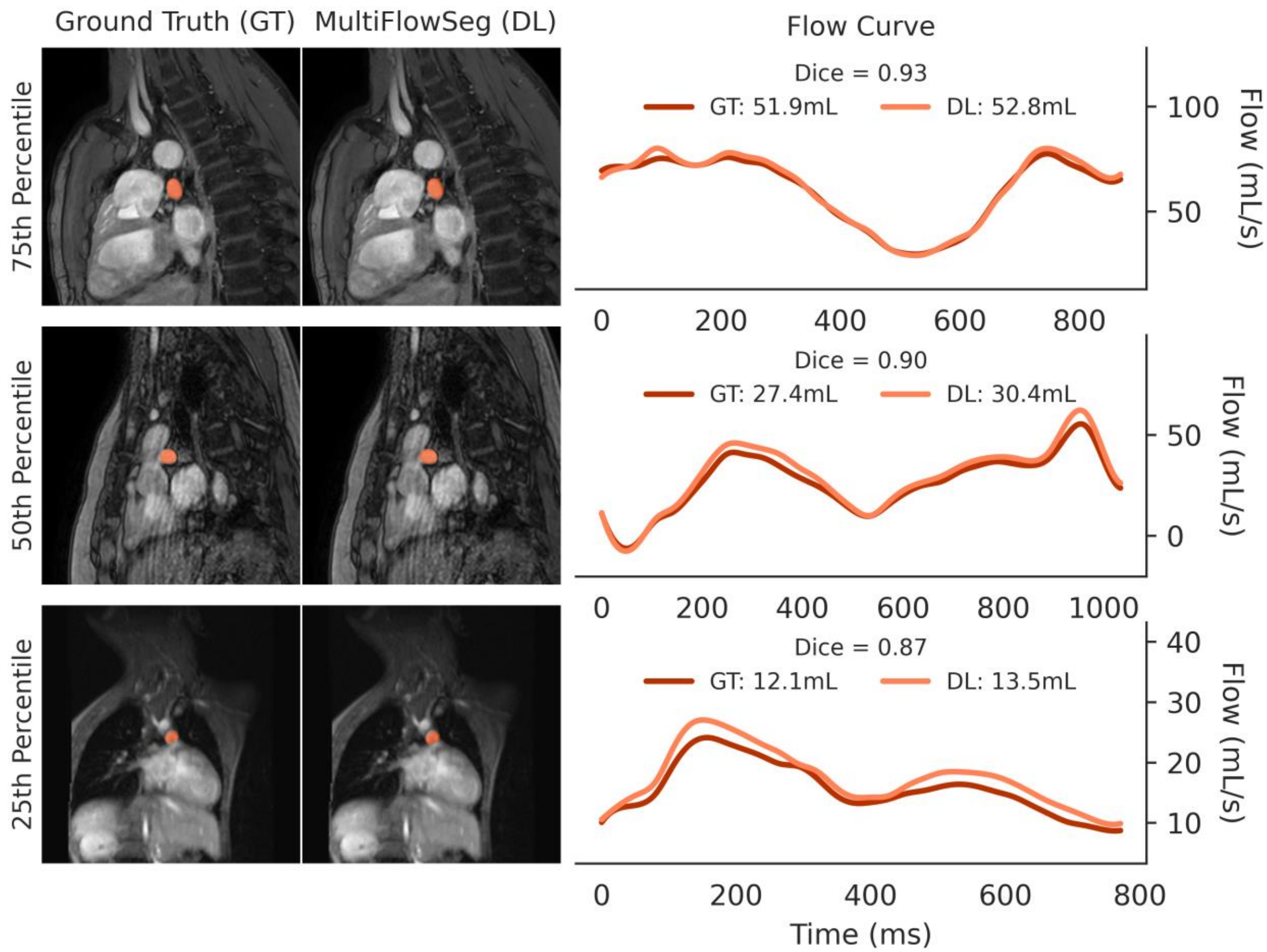

***Supplementary Figure 2: Ground truth and predicted segmentation of the left pulmonary artery for the 25th, 50th, and 75th percentile segmentation accuracy. Corresponding flow curves and net forward volumes are derived from the segmentations.***

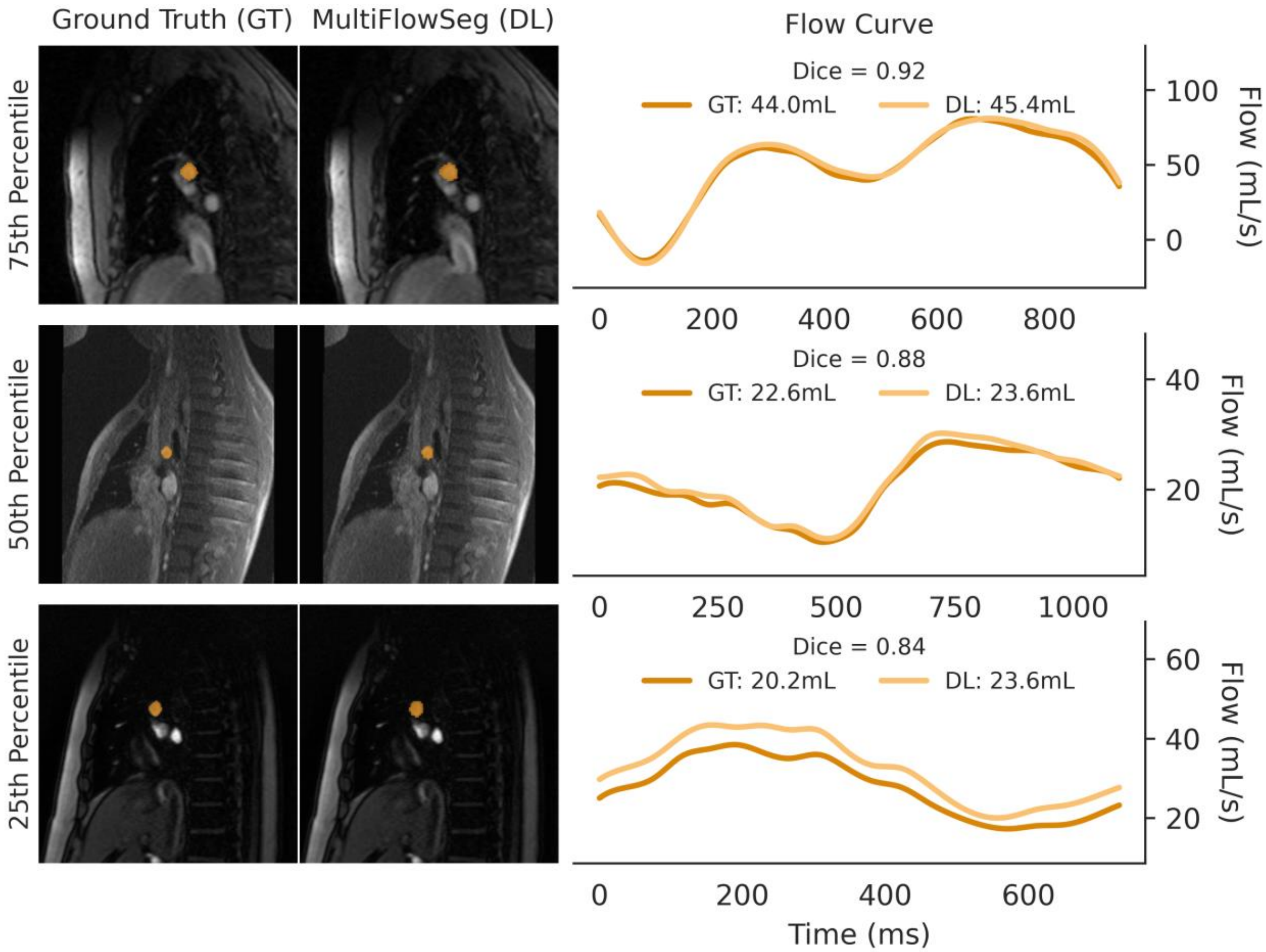

Supplementary Figure 3: Ground truth and predicted segmentation of the right pulmonary artery for the 25th, 50th, and 75th percentile segmentation accuracy. Corresponding flow curves and net forward volumes are derived from the segmentations

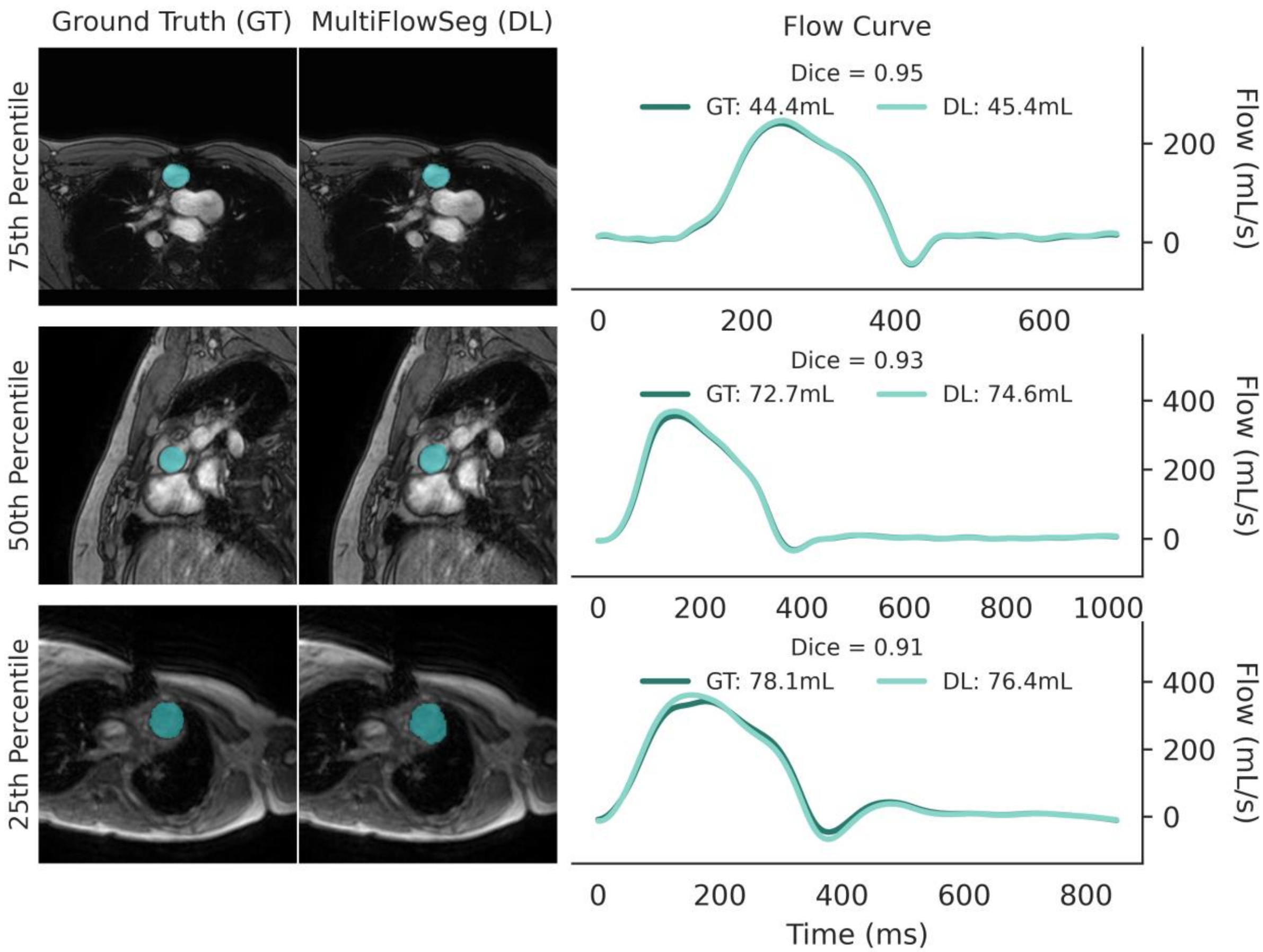

Supplementary Figure 4: Ground truth and predicted segmentation of the aorta for the 25th, 50th, and 75th percentile segmentation accuracy. Corresponding flow curves and net forward volumes are derived from the segmentations

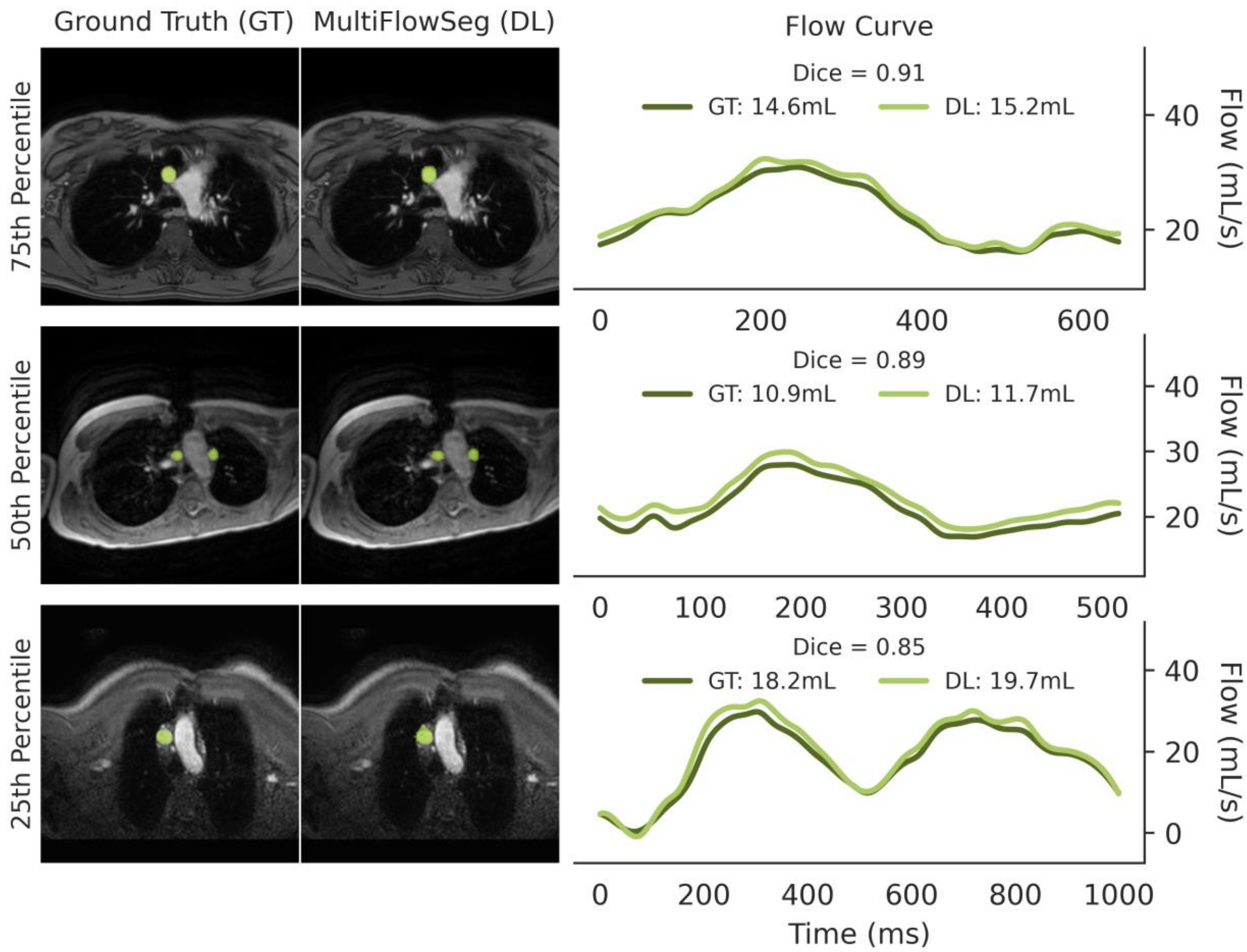

***Supplementary Figure 5: Ground truth and predicted segmentation of the superior vena cava for the 25th, 50th, and 75th percentile segmentation accuracy. Corresponding flow curves and net forward volumes are derived from the segmentations***

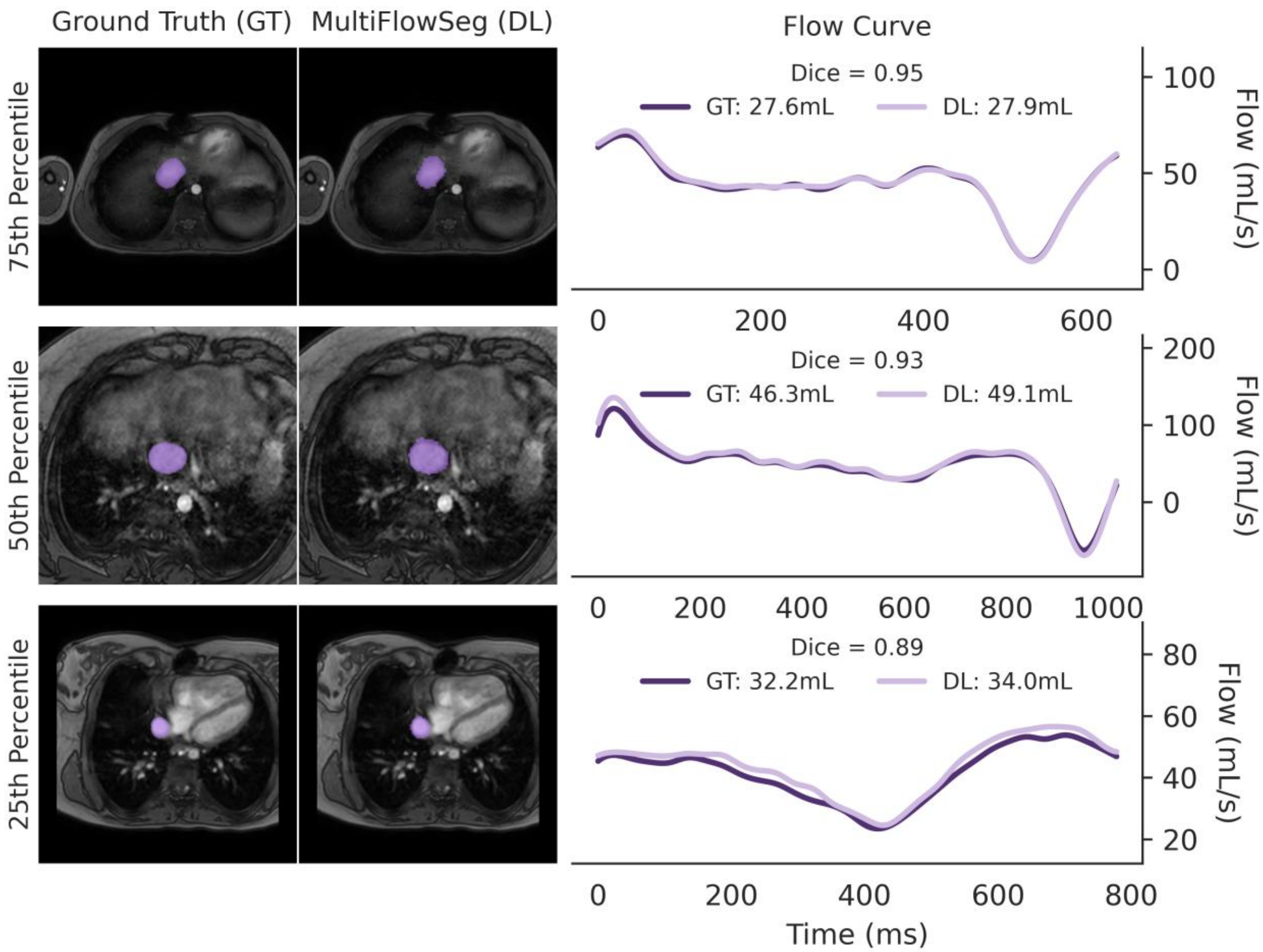

***Supplementary Figure 6: Ground truth and predicted segmentation of the inferior vena cava for the 25th, 50th, and 75th percentile segmentation accuracy. Corresponding flow curves and net forward volumes are derived from the segmentations***

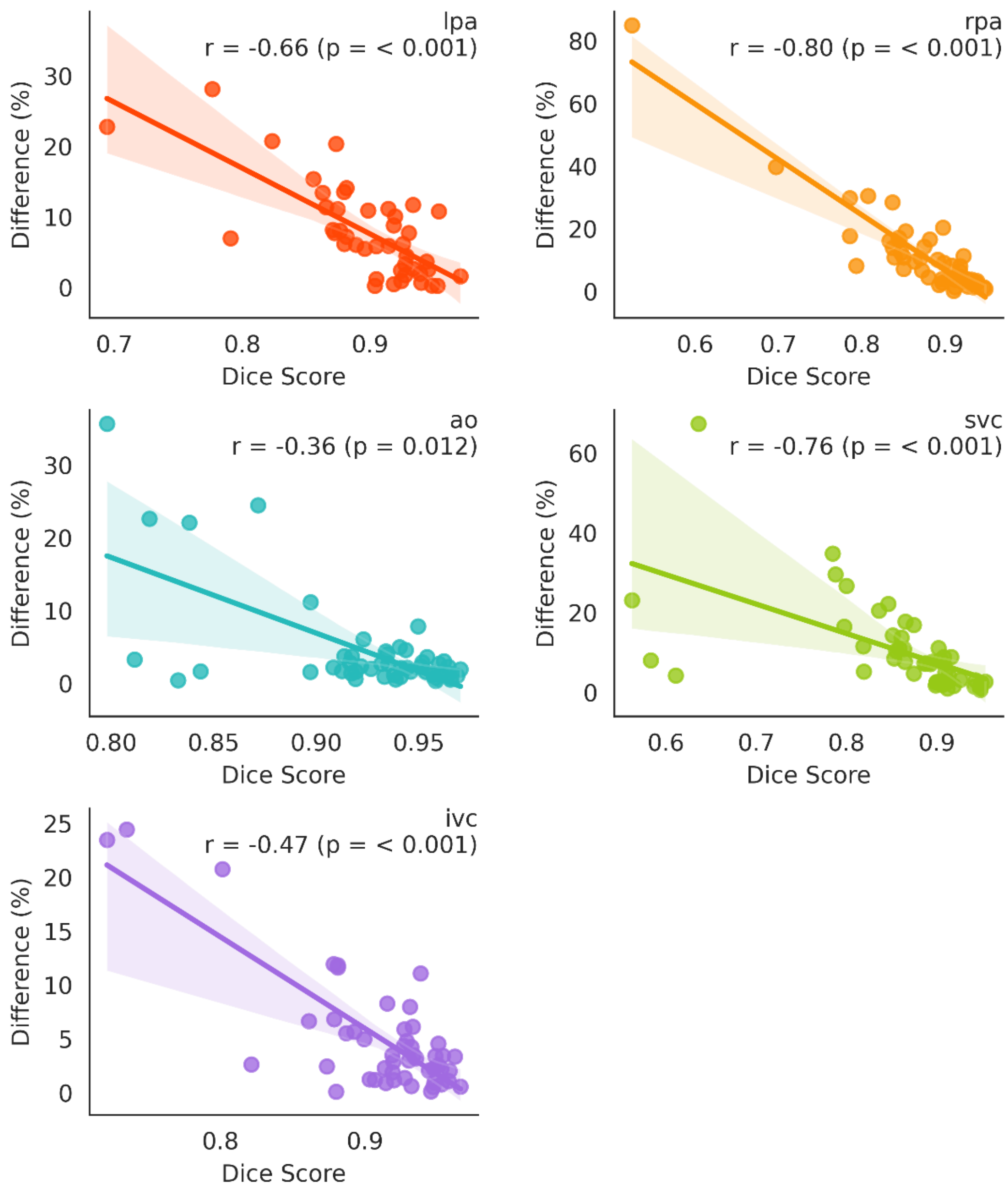

***Supplementary Figure 7. Correlation plot between dice score and difference in net forward volume derivied from DL and manual segmentations as a percentage.***

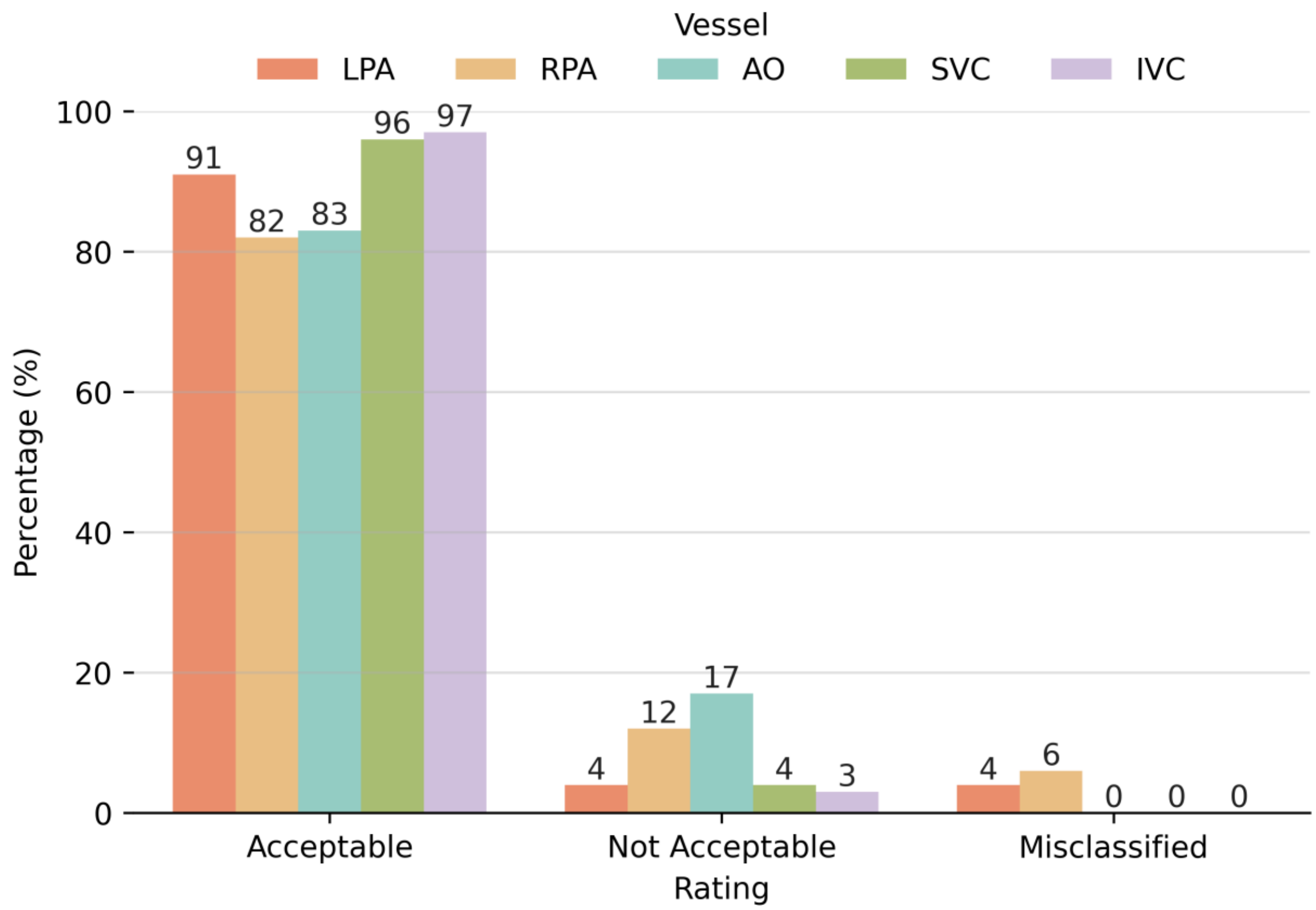

***Supplementary Figure 8. Bar plot of the qualitative review of five vessel segmentations in the pipeline validation dataset (n = 2902 exams; 14510 individual 2D+time segmentations).***

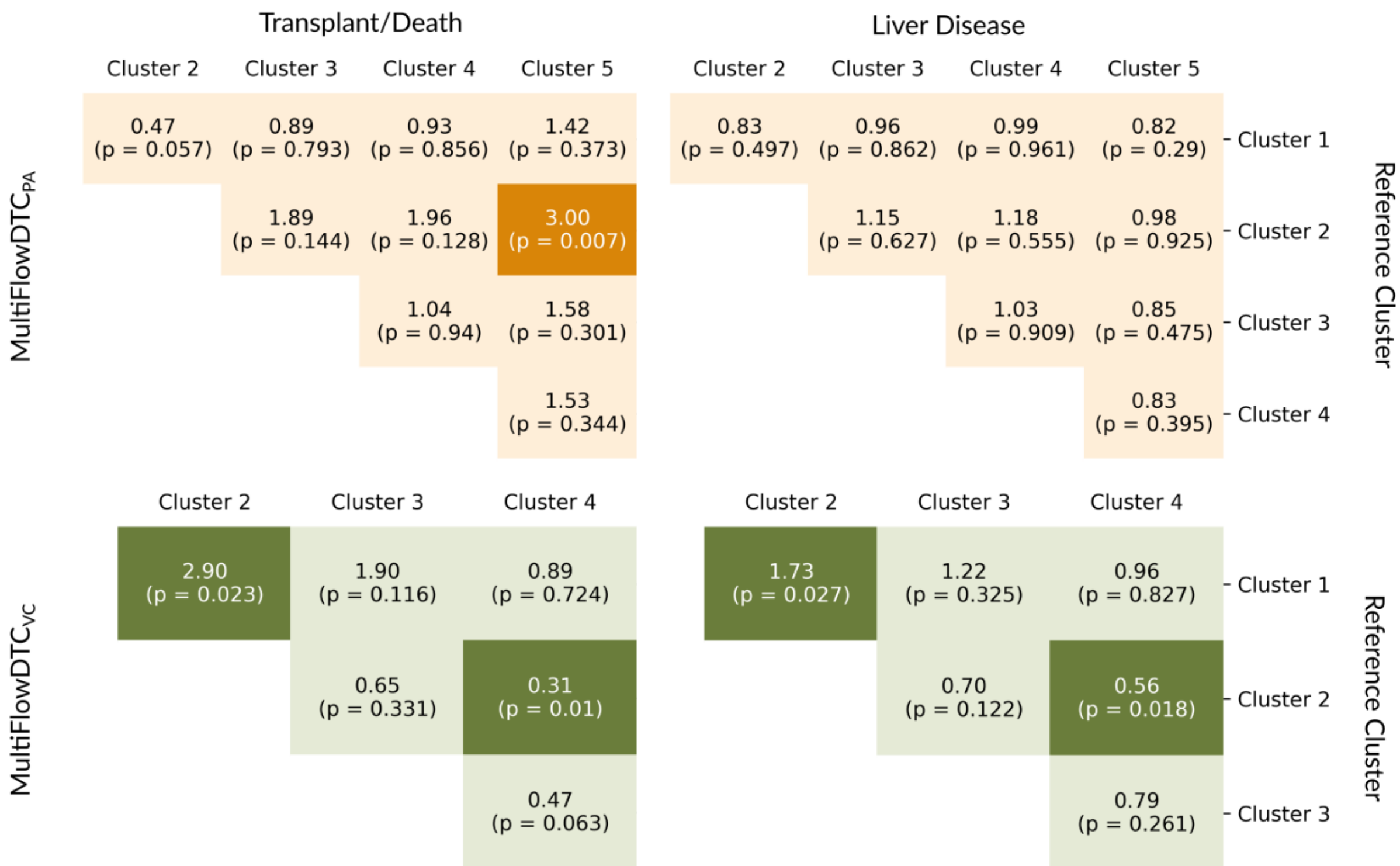

***Supplementary Figure 9. Heatmaps displaying hazard ratios and p-values for pairwise comparisons between clusters derived from both the LPA/RPA and SVC/IVC models of MultiFlowDTC. Hazard ratios for liver disease and transplant/death were calculated using time-varying Cox models and adjusted for age, sex, ejection fraction, and body surface area–indexed aortic flow rate.***

***Supplementary Movie 1. Example input pair of magnitude and imaginary images used in the MultiFlowSeg model. The imaginary component highlights flowing blood with high signal intensity, while static tissue exhibits near-zero signal.***

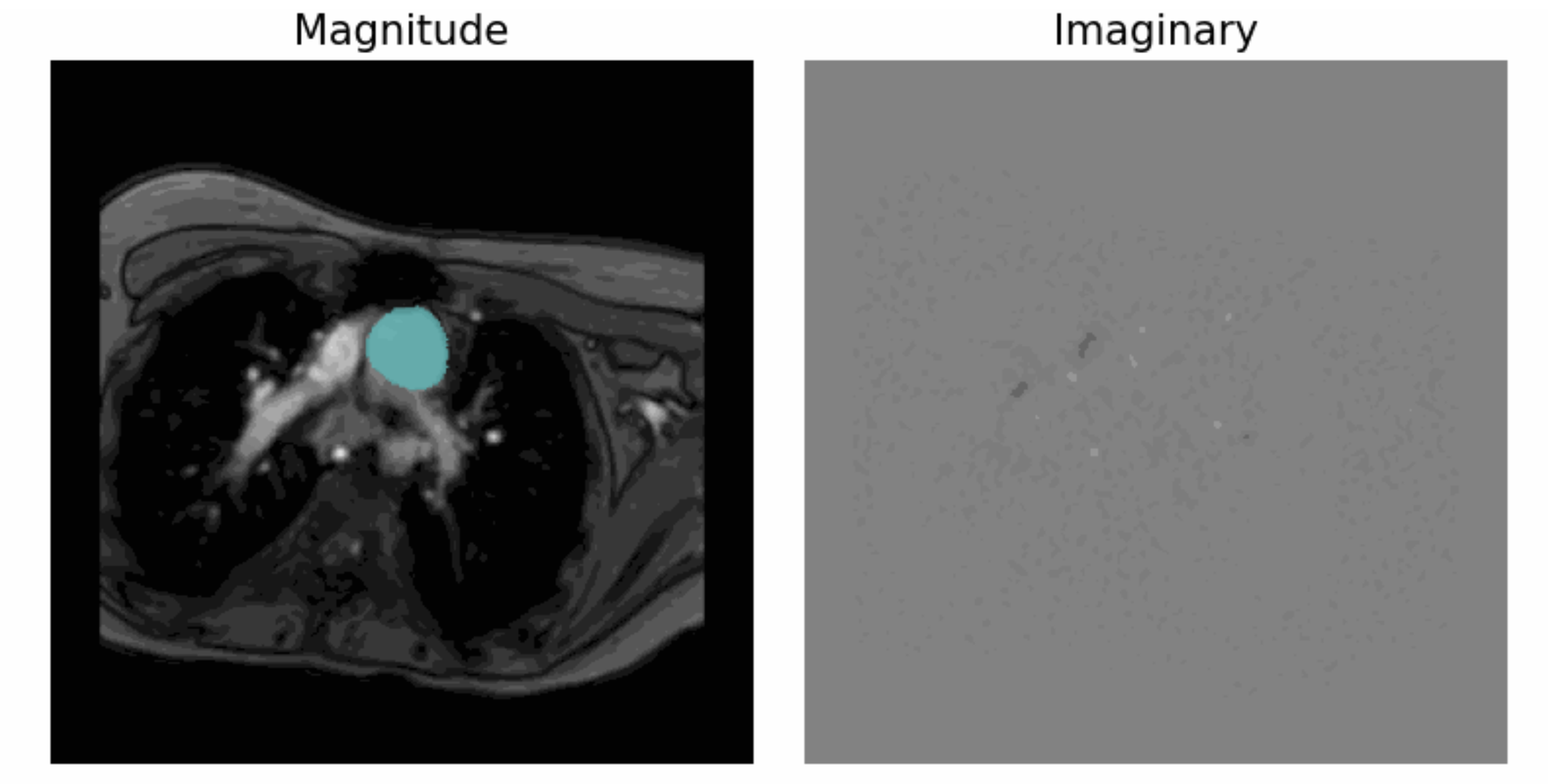